\documentclass[10pt,journal,compsoc]{IEEEtran}
\ifCLASSOPTIONcompsoc
  \usepackage[nocompress]{cite}
\else
  \usepackage{cite}
\fi
\ifCLASSINFOpdf
\else
\fi

\usepackage{epsfig}
\usepackage{graphicx}
\usepackage{amsmath}
\usepackage{amssymb}
\usepackage{float}
\usepackage{xcolor}

\begin{document}
%
\title{Contextual Information Based Anomaly Detection for a Multi-Scene UAV Aerial Videos }
%
%
%
%

\author{Girisha~S $^\dag$,
        Ujjwal~Verma $^\dag$*,~\IEEEmembership{Senior Member,~IEEE,}
        Manohara~Pai M M*,~\IEEEmembership{Senior Member,~IEEE,}
        and~Radhika~M Pai*,~\IEEEmembership{Senior Member,~IEEE}
\thanks{\dag ~ Equal Contribution}
\thanks{* ~ Corresponding Authors}
\IEEEcompsocitemizethanks{\IEEEcompsocthanksitem Girisha S, Manohara Pai M M and Radhika M Pai is with the Department
of Information and communication technology, Manipal Institute of Technology, Manipal Academy of Higher Education, India \protect\\
E-mail: mmm.pai@manipal.edu; ujjwal.verma@manipal.edu;
\IEEEcompsocthanksitem U Verma is with Department of Electronics and Communication Engineering, Manipal Institute of Technology Bengaluru, Manipal Academy of Higher Education, India.}
}

\IEEEtitleabstractindextext{%
\begin{abstract}
UAV based surveillance is gaining much interest worldwide due to its extensive applications in monitoring wildlife, urban planning, disaster management, campus security, etc. These videos are analyzed for strange/odd/anomalous patterns which are essential aspects of surveillance. But manual analysis of these videos is tedious and laborious. Hence, the development of computer-aided systems for the analysis of UAV based surveillance videos is crucial. Despite this interest, in literature, several computer aided systems are developed focusing only on  CCTV based surveillance videos. These methods are designed for single scene scenarios and lack contextual knowledge which is required for multi-scene scenarios.  Furthermore, the lack of standard UAV based anomaly detection datasets limits the development of these systems. In this regard, the present work aims at the development of a Computer Aided Decision support system to analyse UAV based surveillance videos. A new UAV based multi-scene anomaly detection dataset is developed with frame-level annotations for the development of computer aided systems. It holistically uses contextual, temporal and appearance features for accurate detection of anomalies. Furthermore, a new inference strategy is proposed that utilizes few anomalous samples along with normal samples to identify better decision boundaries. The proposed method is extensively evaluated on the UAV based anomaly detection dataset and performed competitively with respect to state-of-the-art methods. 
\end{abstract}

}

\maketitle

\IEEEdisplaynontitleabstractindextext

%
\IEEEpeerreviewmaketitle

\IEEEraisesectionheading{\section{Introduction}\label{sec:introduction}}

\IEEEPARstart{A}{nomalous} events are defined as an occurrence of odd incidents. The detection of these events is crucial for providing security. In general, anomalous events can be broadly grouped into three categories namely: spatial anomalies, temporal anomalies and appearance anomalies \cite{ramachandra2020survey}, \cite{chalapathy2019deep}. Spatial anomalies are related to the location of the object with respect to the scene. Temporal and appearance anomalies are based on the trajectories and appearance of the object respectively.  The definition of anomaly is mostly subjective and depends on the context of the scene \cite{chalapathy2019deep}. An event that is an anomaly in one scene may be normal in another scene. For instance, a truck on road is normal, while on sidewalk is anomaly.  Hence, defining anomaly is challenging. But, accurate and early detection of these events can reduce the risk to human life and thus is an essential aspect of security \cite{santhosh2020anomaly}, \cite{liu2018future}.

\par The current technology available for providing security, analyzes the videos acquired from monitoring systems such as CCTVs installed at fixed locations. These surveillance videos are manually analysed by security personal for detecting anomalies which is time-consuming and tedious process. Therefore, the past few years have seen several works on designing algorithms for detecting anomalous activities in videos \cite{Ionescu_2019_CVPR}, \cite{xu2015learning}, \cite{ma2015anomaly}. However, a majority of existing methods analyse videos from static surveillance cameras installed at fixed locations such as pedestrian walkaway (CUHK Avenue \cite{lu2013abnormal}, Ped1, Ped2 \cite{mahadevan2010anomaly}) Subway entrance/exit \cite{adam2008robust}.  In the recent past, a few studies have focused on analysing videos from multiple scenes \cite{liu2018future}. However, the videos analysed in these works contain either minor or no camera motion. 
There is hardly any study on detecting anomalous activity in a video acquired by a moving camera. Also, there are very few standard datasets available for the development of multi-scene anomaly detection algorithms.

\begin{figure*}[t]
	\includegraphics[width=7.2in,height=2.7in]{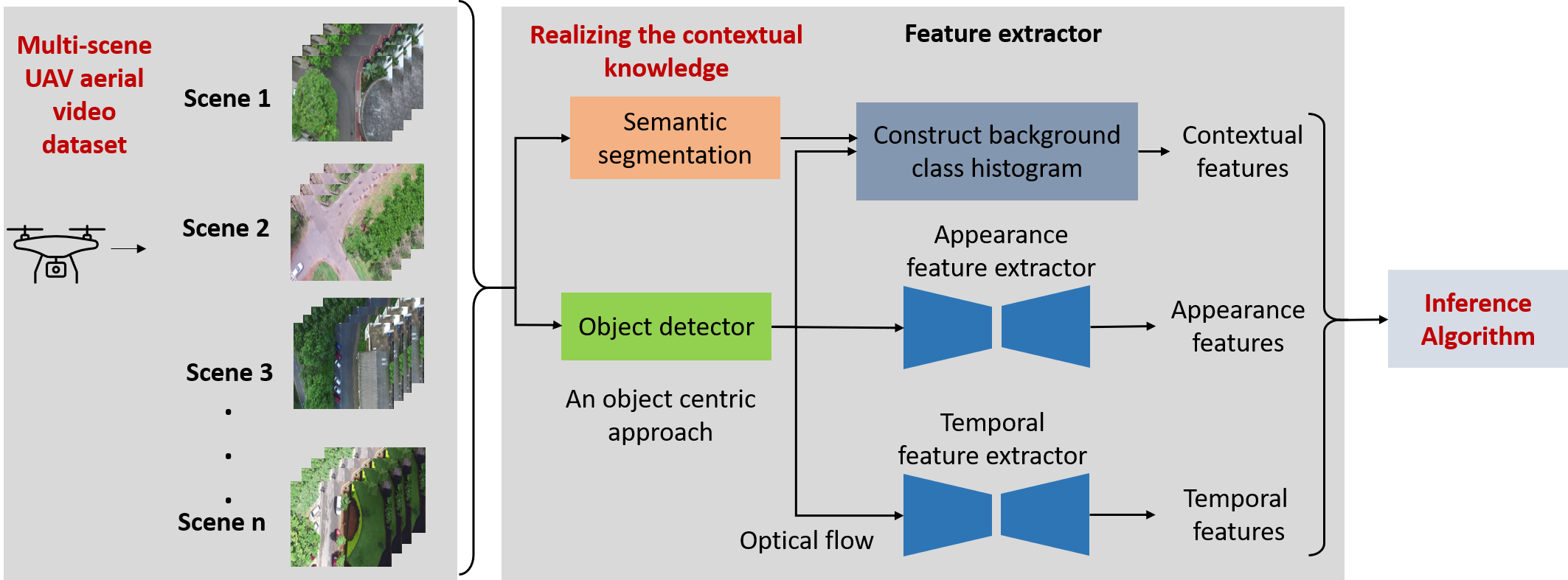}
	\caption{Overview of proposed model: Multi-scene UAV aerial video dataset creation. Object-centric contextual, appearance and temporal features are first extracted from each frame of the aerial video. The inference algorithm is formulated as a multi-class classification algorithm on these extracted features to identify anomalous events. }
	\label{overall}
\end{figure*}

Last few years have seen an increased interest in using Unmanned Aerial Vehicles (UAVs) for surveillance \cite{teng2021viewpoint}. Compared to a static CCTVs with a fixed coverage area, videos acquired from UAV based surveillance systems can cover larger area with varying perspectives. Besides, UAVs have the advantage of mobility and can be rapidly deployed. Despite the better flexibility and mobility offered by UAV based surveillance system, there are limited works on anomaly detection in videos acquired using UAVs \cite{chriki2021deep},\cite{9219585}, \cite{9341790}. However, the videos for these studies are acquired at a single location (car parking) and the variations in the background scene information are limited. The present study focuses on detecting anomalous activities by analysing videos from UAVs. These videos are acquired at multiple locations (multi-scene) and consist of significant camera motion (Figure \ref{overall}) with varying viewpoints. No published reports are available on this kind of multi-scene anomalous activity detection from UAV aerial videos with significant variations in background scene information.

In general, an anomaly detection algorithm comprises of two steps namely training and detection \cite{Ramachandra2020}, \cite{Ionescu_2019_CVPR}\cite{doshi2020any}. The training phase consists of developing a model for normal activities using the features extracted from the training videos containing only normal activities. In the detection phase, an anomaly score is assigned to the test video by feeding the same types of extracted features to the developed model. 
However, these models are only trained on normal patterns and hence the decision boundary may not accurately discriminate anomalous patterns.
There exist other methods which address the problem of anomaly detection as an outlier detection where all the outliers are considered anomalous \cite{xu2017detecting}, \cite{zhao2011online}, \cite{cheng2015video}.  However, these methods fail to detect local anomalies which can be defined as those patterns that are closer to normal patterns but are anomalous in nature.  For example, consider a situation of a stationary vehicle present in a parking zone vs a stationary vehicle present on road. Here, the appearance and temporal characteristics of both vehicles are similar. However, the vehicle present on road is an example of a local anomaly since its characteristics are similar to normal patterns.   Recently, few studies focuses on features extracted from object of interest (pedestrian). This object centric approach allows to accurately locate the anomalous object in the frame \cite{Ionescu_2019_CVPR}.

The majority of the anomalous activity detection algorithms are developed for single-scene scenarios where the background is constant. Hence, these models cannot be adopted to detect anomalies in multi-scene scenarios such as UAVs where the background changes and hence the context of the scene. In these situations, realizing the context of the scene is important as it facilitates accurate detection of anomalies. For instance, a pedestrian on the road is an anomalous event while a person walking on sidewalk is a normal event. Therefore, the background scene information (road/sidewalk) would be useful in detecting an anomalous event. This work proposes to integrate the background scene information along with appearance and motion information for anomalous activity detection. To the best of our knowledge, there is no existing work on incorporating background scene information for anomalous activity detection.

\par In the present study, an object-centric approach is presented which extracts   contextual (background scene), appearance and motion features with a focus on object of interest (vehicles and human). A four class semantic segmentation algorithm is utilized to extract the background scene information, while autoencoder based reconstruction error is utilized for extracting appearance and motion features. Subsequently, an inference algorithm which constitutes a set of one class SVM classifiers, are trained on these features to detect anomalies. This study proposes a new inference algorithm to handle local anomalies. We argue that incorporating features extracted from few anomalous  videos can improve the performance of the inference model. The contributions of the research are as follows:
\begin{enumerate}
	\item A new multi-scene UAV anomaly detection dataset (MUAAD: Manipal UAV Anomalous Activity Dataset) is presented. This provides a common platform for researchers to compare the developed models for UAV-based anomaly detection.
	\item A new object-centric method for anomaly detection is proposed which captures contextual, temporal and appearance-based features. This allows the model to process multi-scene scenarios and detect spatial, temporal and appearance based anomalies.
	\item A new inference algorithm is introduced to detect anomalous patterns. This algorithm is trained such that  it is capable of detecting local anomalies efficiently which in-turn improves the accuracy of detection. 
\end{enumerate}
The structure of the paper is as follows: Section \ref{lite} presents the recent literature regarding the anomaly detection in videos. The details of the proposed dataset is presented in Section \ref{dataseta}. Section \ref{met} describes the methodology for detecting anomalies from UAV aerial videos. Section \ref{res} presents the various results of the proposed system. Finally, conclusion of the paper is given in section \ref{con}.

\section{Related works}
\label{lite}
\par Various approaches have been put forward to solve the issue of anomaly detection in surveillance videos. Also, several datasets are available for the development and evaluation of anomaly detection models. A detailed discussion of these can be found in \cite{ramachandra2020survey}, \cite{santhosh2020anomaly}, \cite{chalapathy2019deep}. In this section, initially, A summary of various datasets available for anomaly detection followed by a brief discussion on various methods proposed for video anomaly detection has been presented.

\subsection{Video anomaly detection datasets}
\par The development of anomaly detection system for videos is dependent on the availability of annotated datasets. Furthermore, they provide a common platform for the researcher to evaluate the developed algorithm.  In literature, several popular benchmark datasets are available for anomaly detection \cite{adam2008robust}, \cite{raghavendra2006unusual}, \cite{mahadevan2010anomaly}, \cite{lu2013abnormal}. The Subway dataset proposed in \cite{adam2008robust}, provides videos collected at a subway entrance and exit. The UMN \cite{raghavendra2006unusual} dataset has 11 outdoor video scenes with staged anomalies. In \cite{lu2013abnormal}, the authors proposed CUHK Avenue dataset which has 37 videos collected from a single camera which contains  47 anomalous events. UCSD ped1 and ped2 \cite{mahadevan2010anomaly} are another popular dataset used for anomaly detection. Ped 1 and Ped 2 dataset contains video acquired from two stationary camera overlooking pedestrian walkaways. The videos of UCSD ped1 and ped2 are collected at a low resolution. All these datasets are developed for single-scene video anomaly detection. A few multi-scene video anomaly detection datasets developed are available in literature such as ShangaiTech \cite{liu2018future} and UCF crime \cite{sultani2018real}. ShangaiTech dataset contains videos collected from 13 different scenes with 130 abnormal events. However, all these datasets have videos collected from static CCTV cameras with a limited coverage area and fixed background. In the present study, a new anomaly detection dataset containing the videos acquired from the camera on-board a moving Unmanned Aerial Vehicle (UAV) is proposed. Unlike existing datasets, the proposed dataset contains background scene information in the form of pixel level mask for four classes (greenery, road, construction and water bodies).  The contextual scene information can be inferred from these masks which would aid in identifying anomalous activities in multi-scene videos with significant variation in the background information.

\subsection{Video anomaly detection}
\par Video anomaly detection algorithms can be broadly grouped into three categories namely: distance-based methods, probabilistic methods and reconstruction based methods. The distance-based methods construct a model of normal patterns \cite{Ramachandra_2020_WACV}, \cite{Ionescu_2019_CVPR}, \cite{Sabokrou_2018_CVPR}, \cite{ionescu2019detecting}, \cite{ravanbakhsh2018plug}. The distance of test patterns from the model is utilized to estimate the anomaly score. In these methods, handcrafted features \cite{dalal2006human}, \cite{dalal2005histograms}, as well as deep features \cite{ravanbakhsh2018plug}, \cite{sabokrou2018deep}, \cite{xu2015learning}, are widely used to represent the appearance and temporal domain of the object. Subsequently, one class SVMs are popularly used in these methods to identify the decision boundaries of normal patterns required for estimating distances \cite{Ionescu_2019_CVPR}, \cite{xu2015learning}, \cite{ma2015anomaly}. Also, distance metrics such as Mahalanobis is used to measure the distance from normal patterns \cite{sabokrou2018deep}, \cite{sabokrou2017deep}. The probabilistic methods measure anomaly scores in probability space \cite{antic2015spatio}, \cite{kim2009observe}, \cite{benezeth2009abnormal}, \cite{kratz2009anomaly}, \cite{feng2017learning}, \cite{cheng2015video}. Gaussian Mixture Models and Markov Random Fields are popularly used traditional methods \cite{kratz2009anomaly}, \cite{kim2009observe}.  In the reconstruction based method, the given input data is converted to a high-level representation or a latent space \cite{Ionescu_2019_CVPR}, \cite{vu2019robust}, \cite{chong2017abnormal}, \cite{hasan2016learning}. Further, these inputs are reconstructed from this high-level representation. Since these models are trained only on normal patterns, any anomalies would produce higher reconstruction errors which are used for determination of anomalies. To this end, auto-encoders are extensively used to transform input to latent space. Recently, GAN based methods are gaining popularity for anomaly detection. However, these methods do not consider the contextual knowledge required for detecting multi-scene video anomalies.

The anomalous activities occur rarely and developing a robust model for detection of anomalous activities is a challenging task. Hence, popular methods proposed in literature are semi-supervised where the model is trained on normal patterns. However, these models fails to detect local anomalies. In this context, this work formulates the inference algorithm such that it utilizes a few samples of anomalous events to identify better decision boundaries. Note that a recent work \cite{doshi2020any} formulated the few shot anomaly detection task as identifying the anomalous activity with few sequential frames of a video. However,  our proposed approach utilizes the features from few anomalous videos to learn a more robust model. To the best of our knowledge,  very limited work exists that considers minor supervision to anomalous samples to improve the performance of anomaly detection in UAV videos. 

\par The scarcity and ambiguous definition of anomalies make it challenging to develop anomaly detection datasets. Also, lack of standard UAV based anomaly detection datasets, make it difficult to develop Computer Aided Decision (CAD) support systems. Hence, it urges to develop a new dataset for detecting anomalous activities in UAV aerial videos. Besides, the existing methods for anomalous activity detection are designed for single-scene videos with a fixed or constant background. Therefore, these approach would have limited success when there is a significant change in the background scene information, such as the video from the camera on-board a moving UAV.  The UAV videos contain a different perspective view (topdown) of the scene as compared to the videos acquired from CCTVs.   In this regard, the present work aims at the development of CAD systems to detect anomalies from UAV aerial videos. Furthermore, this study presents a novel feature extractor that holistically extracts contextual, temporal and appearance features required for multi-scene video anomaly detection. Also, the present work proposes a new inference algorithm to efficiently determine the decision boundaries needed for the accurate detection of anomalies.  
\begin{table*}[t]
\centering	
	\caption{Comparison of MUAAD with other datasets. }
	\label{datasetc}
\begin{tabular}{|c||c|c|c|c|c|c|}
	\hline
	
	\textbf{Dataset} & \textbf{Total No. Of Videos}& \textbf{Annotated frames}&\textbf{Anomalous events}& \textbf{Multi-scene}& \textbf{Camera motion}&\textbf{Modality} \\
    
	\hline
	CHUK Avenue & 37 &30,652&47& No &Limited camera shake&CCTV\\
	\hline
	ShanghaiTech &437 &317,398 & 130&\checkmark&-&CCTV\\
	\hline
	UCSD Ped1 &70&14,000&40&-&-&CCTV\\
	\hline
	UCSD Ped2 &28&4,560&12&-&-&CCTV\\
	\hline
	MUAAD (Ours) &60&68,687&92&\checkmark&\checkmark&UAV\\
	\hline
	
\end{tabular}
\end{table*}

\begin{table}[t]
\centering	
	\caption{Details of the proposed dataset.}
	\label{dataset}
\begin{tabular}{|c||c|}
	\hline
	
	\textbf{Training videos} & 8 \\
	\hline
	\textbf{Testing videos} &52 \\
	\hline
	\textbf{Number of scenes} & 9 \\
	\hline
	\textbf{Objects considered} & Vehicle, Humans \\
	\hline
	\textbf{Number of annotated frames} & 68,687 \\
	\hline
	\textbf{Types of anomalous events} & 7 \\
	\hline
	\textbf{Occurrence of anomalous events} & 92 \\
	\hline
	\textbf{Altitude} & 25-30 mts\\
	\hline
	\textbf{Resolution} & 1280x720p\\
	\hline
	\textbf{Frame rate} &29.97 frames/second \\
	\hline
	\textbf{Minimum duration} & 7 seconds \\
	\hline
	\textbf{Maximum duration} & 1 min 30 seconds\\
	\hline
	\textbf{Camera motion} & Present\\
	\hline
	\textbf{Illumination variation} & Present\\
	\hline
	
\end{tabular}
\end{table}
\begin{figure}[t]
	\begin{tabular}{cc}
		\begin{minipage}{32pt}
			\includegraphics[width=1.6in, height=1.2in]{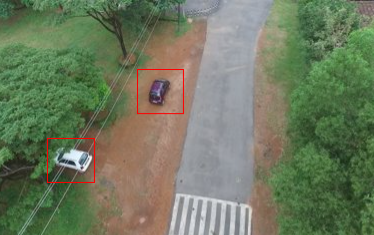}
			\centering{(a)}
		\end{minipage}
		&
		\hspace{2.7cm}
		\begin{minipage}{32pt}
			\includegraphics[width=1.6in, height=1.2in]{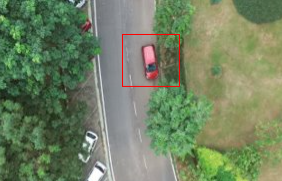}
			\centering{(b) }
		\end{minipage}
	\\
	\\
	
		\begin{minipage}{32pt}
			\includegraphics[width=1.6in, height=1.2in]{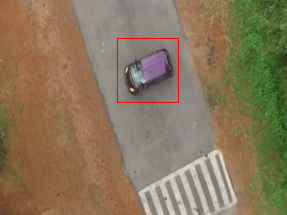}
			\centering{(c)}
		\end{minipage}
		&
			\hspace{2.7cm}
		\begin{minipage}{32pt}
			\includegraphics[width=1.6in, height=1.2in]{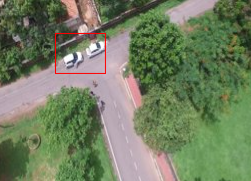}
			\centering{(d)}
		\end{minipage}
		\\
\\
		
	\end{tabular}
	\caption{Few frames from the dataset which contain anomalous event. The anomalous object is shown as a red bounding box. An example of contextual anomaly is shown in (a), while an example of vehicle parked on road is shown in (b). Figure (c) shows an example of vehicle with random trajectories, and (d) shows an example of vehicle parked in no-parking zone. }
	\label{dataset_images}
\end{figure}

\begin{figure}[t]
	\begin{tabular}{cccc}
		\begin{minipage}{32pt}
			\includegraphics[width=0.8in, height=0.8in]{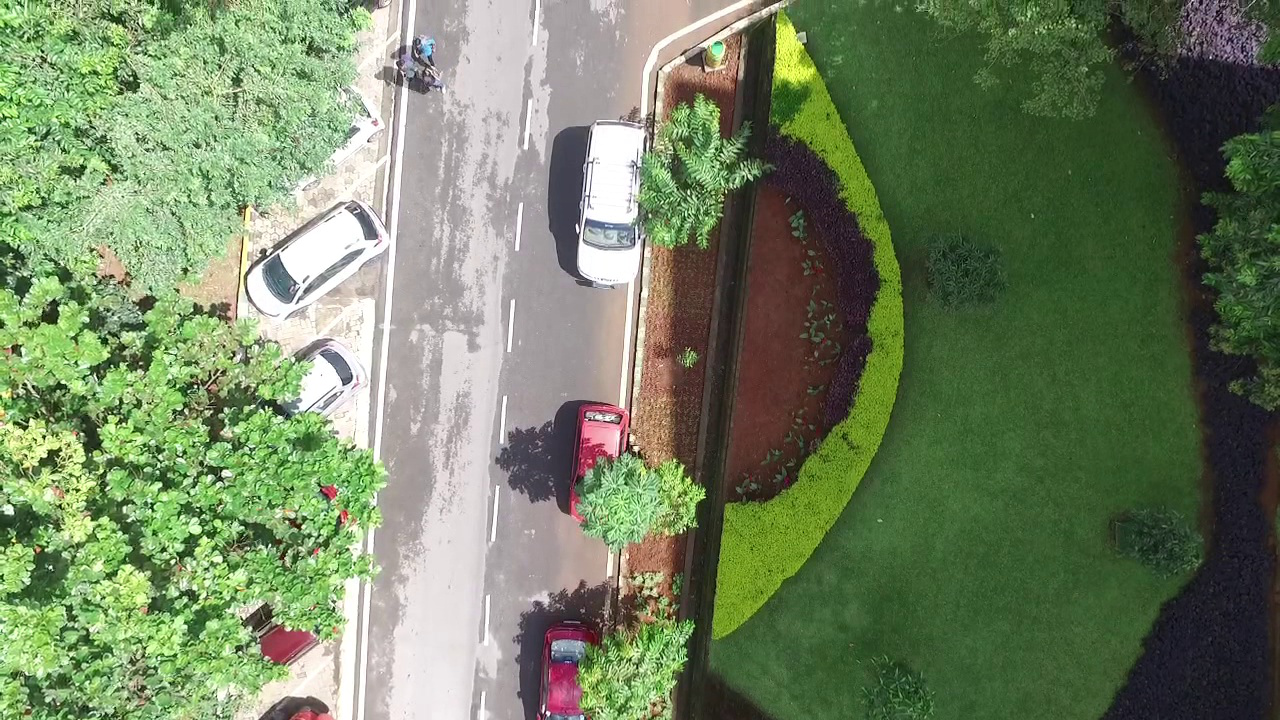}
		\end{minipage}
		&
		\hspace{0.5cm}
		\begin{minipage}{32pt}
			\includegraphics[width=0.8in, 	height=0.8in]{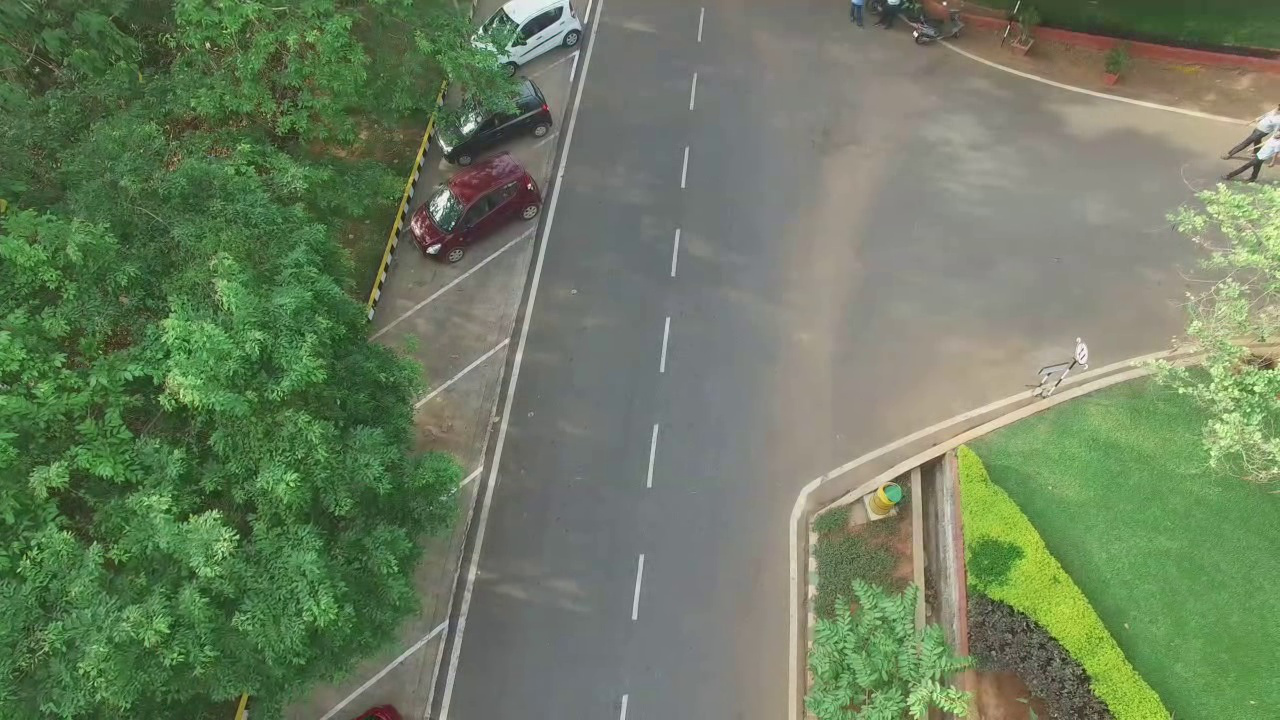}
		\end{minipage}
		&
		\hspace{0.5cm}
		\begin{minipage}{32pt}
			\includegraphics[width=0.8in, 	height=0.8in]{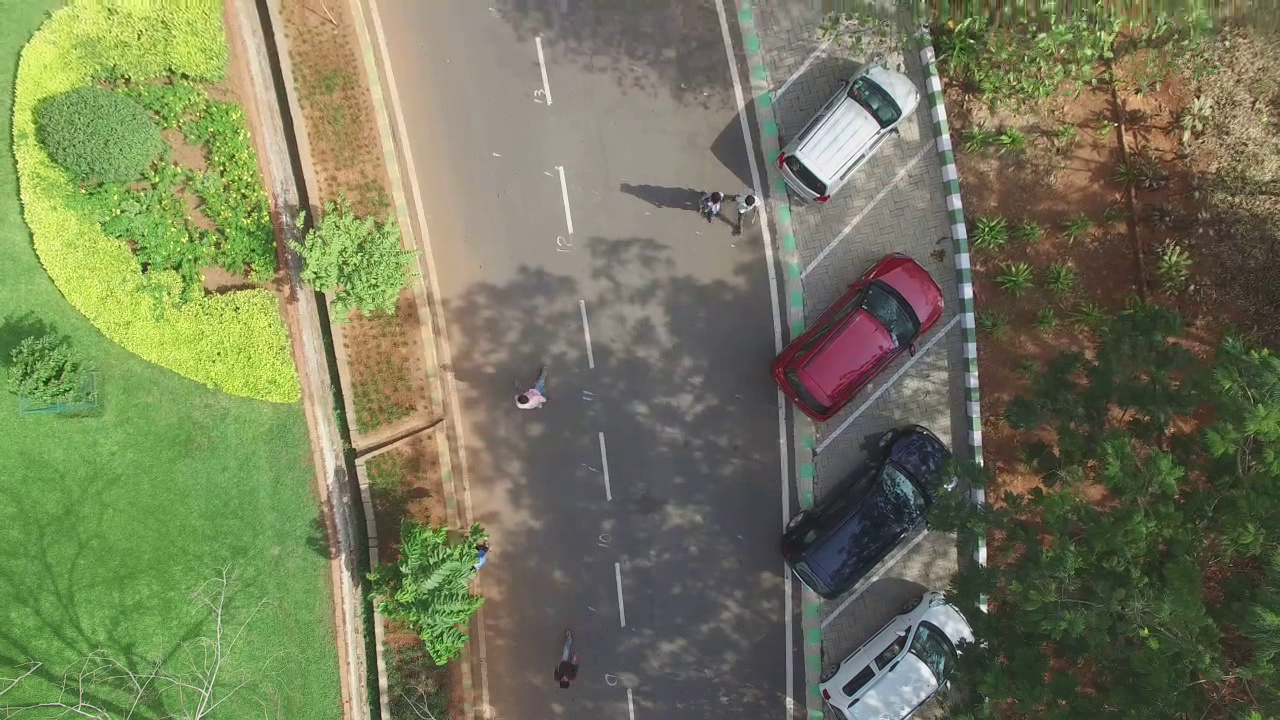}
		\end{minipage}
		&
			\hspace{0.5cm}
		\begin{minipage}{32pt}
			\includegraphics[width=0.8in, 	height=0.8in]{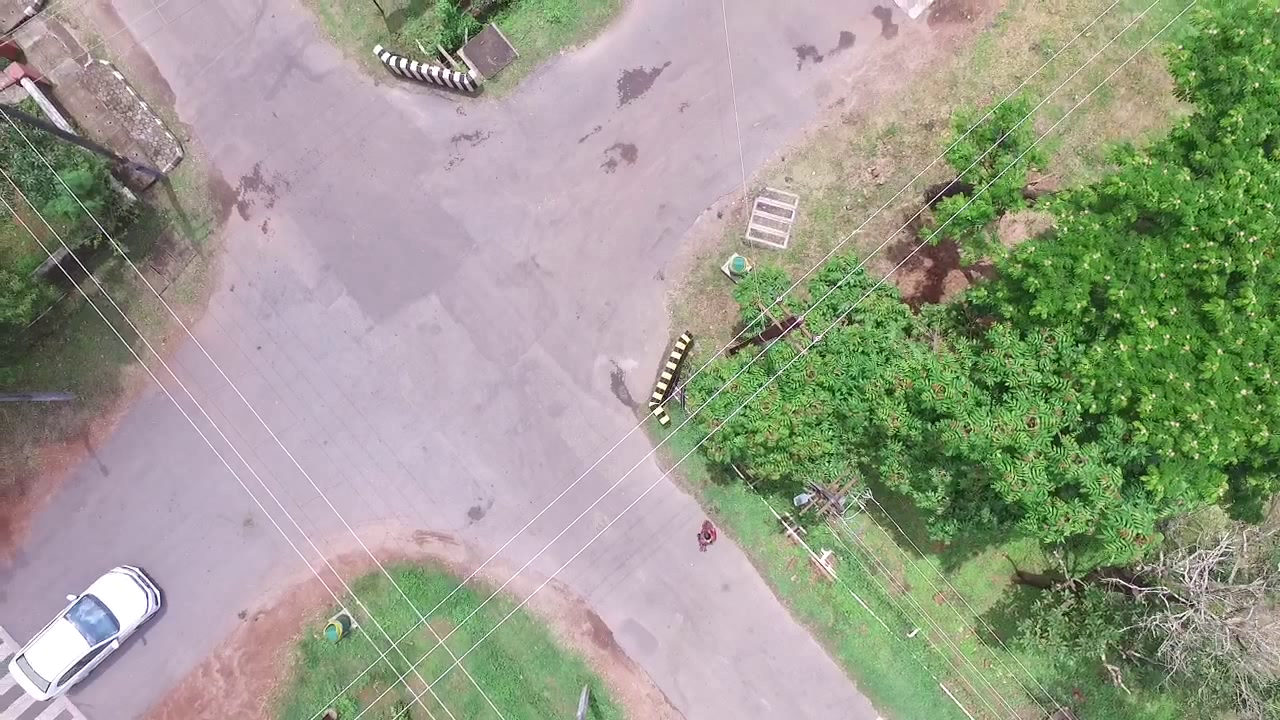}
		\end{minipage}
		\\
\\
		
		\begin{minipage}{32pt}
			\includegraphics[width=0.8in, height=0.8in]{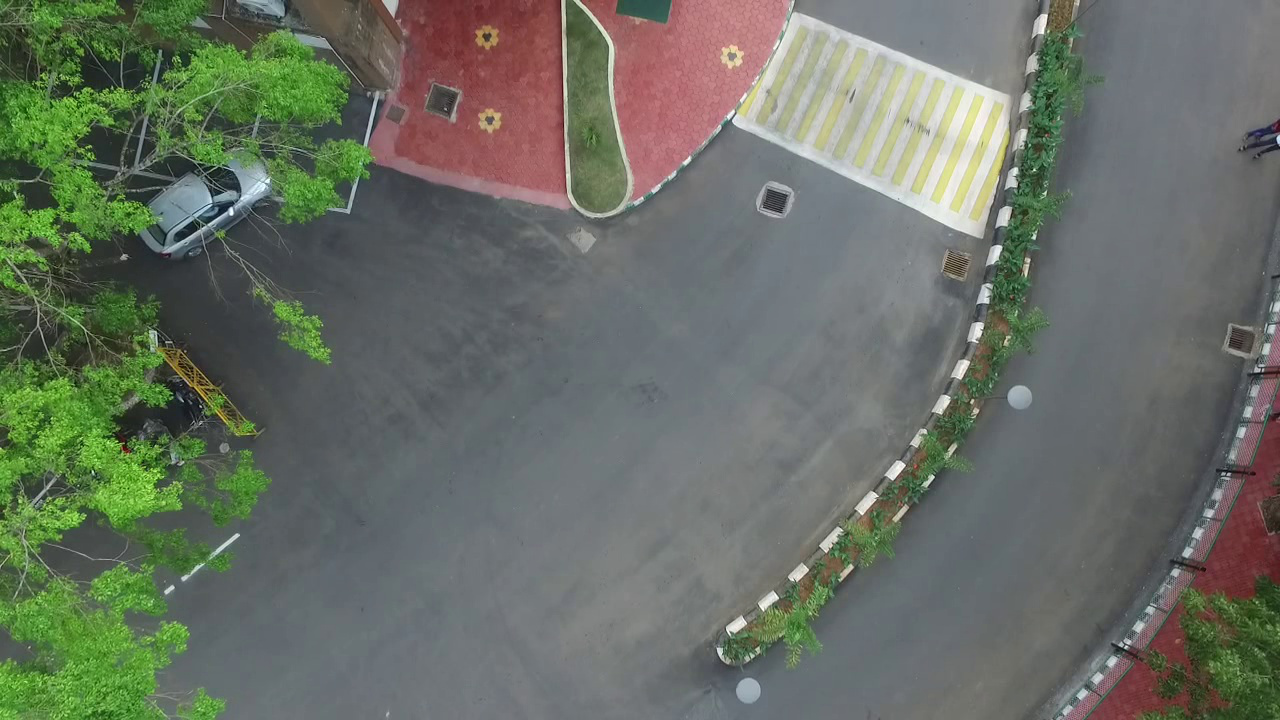}
		\end{minipage}
		&
		\hspace{0.5cm}
		\begin{minipage}{32pt}
			\includegraphics[width=0.8in, 	height=0.8in]{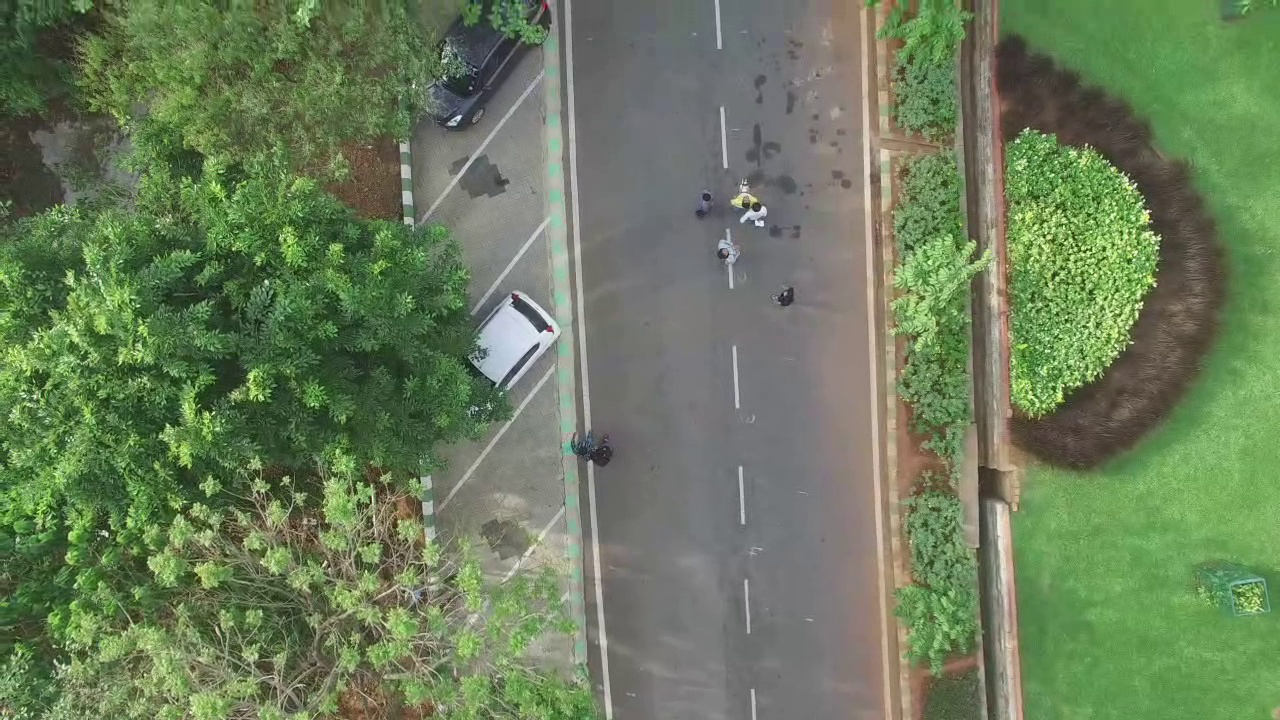}
		\end{minipage}
		&
		\hspace{0.5cm}
		\begin{minipage}{32pt}
			\includegraphics[width=0.8in, 	height=0.8in]{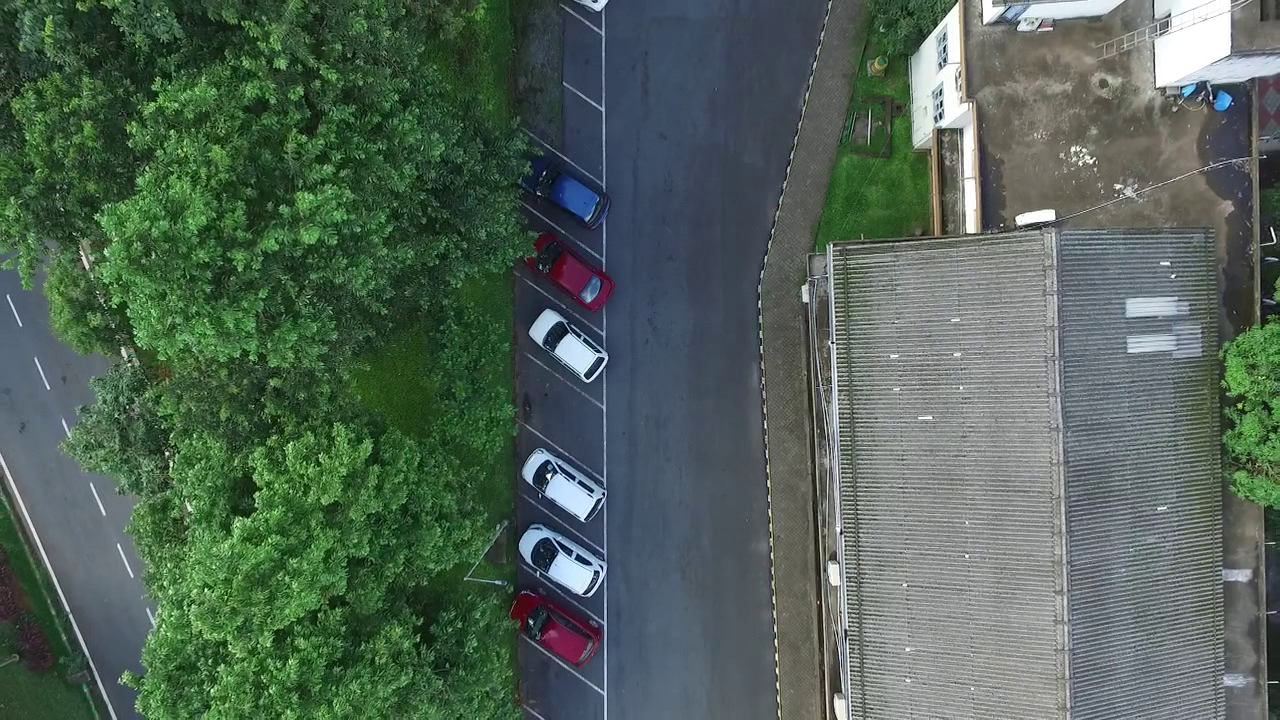}
		\end{minipage}
		&
			\hspace{0.5cm}
		\begin{minipage}{32pt}
			\includegraphics[width=0.8in, 	height=0.8in]{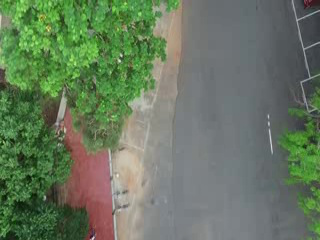}
		\end{minipage}
		
		\\
		\\
		
	\end{tabular}
	\caption{Shows a few sample variations in scene of the proposed UAV anomaly detection dataset.}
	\label{dataset_images1}
\end{figure}

\section{UAV anomaly detection dataset generation}
\label{dataseta}

\par The present study proposes a new UAV aerial video dataset for anomaly detection. The videos are acquired from the camera on-board a UAV (DJI Phantom 3 drone) at the campus of Manipal Institute of Technology, Manipal, India. We refer to this dataset as Manipal UAV Anomalous Activity Dataset (MUAAD). These videos are acquired at 29 fps and are of resolution 1024x720 pixels. The videos are collected during different time of the day at which maximum vehicular traffic is expected (morning 9am, afternoon 12 noon and evening 4pm). The minimum and maximum duration of the videos are 7 seconds and 1 minute 30 seconds respectively. The videos are collected at an altitude of 25-30 metres. In total, the dataset contains 60 videos collected from nine different locations within the campus. 

In the present study, seven general anomalous patterns are considered for annotations, viz. : vehicles parked in a no-parking zone, over-speeding vehicles, frantic trajectories of vehicles, vehicles found on non-pavement areas, pedestrians walking on the road, pedestrians gathering on-road and frantic movements of pedestrians. A few examples of these patterns are highlighted and shown in Figure \ref{dataset_images}. A frame level annotation (normal/anomalous) for each frame in the video is provided for a total of 68,687 frames. All the anomalous frames are labeled as $1$ and normal frames are labeled as $0$. The dataset contains a total of 92 anomalous events distributed in the $60$ videos.  A summary of the proposed dataset is given in Table \ref{dataset}. The training and the testing set consists of 8 and 52 videos respectively. The dataset is available at \emph{https://github.com/uverma/MUAAD}

\par The existing anomaly detection datasets contains videos acquired from static CCTV camera at fixed locations \cite{adam2008robust}, \cite{raghavendra2006unusual}, \cite{mahadevan2010anomaly}, \cite{lu2013abnormal}. The videos present in these dataset contain very little or no camera motion. Beside, there is no  variation in the background scene information in these videos. As discussed in \cite{ramachandra2020survey}, there is a need to develop anomalous activity detection algorithms in multi-scene videos with significant camera motion. The present study attempts to address this research gap by proposing a new multi-scene moving camera dataset. This dataset (MUAAD) contains videos acquired from the camera on-board a moving UAV and collected at 9 distinct \textit{locations} (multi-scene).  A few frames from this dataset is shown in Figure \ref{dataset_images1}. It can be observed that there is significant variation in the background information along with the camera motion. Moreover, the videos acquired from UAV contains a different perspective (topdown) view of the object (vehicles/human) as compared to the front view captured by the CCTVs. 

\section{Methodology}
\label{met}

\subsection{Proposed model}
\par This study proposes an object-centric multi-scene video anomaly detection algorithm for UAV surveillance videos. Figure \ref{overall} shows the overview of the proposed UAV based anomaly detection system. In this work, an object-centric approach is adopted, since anomalies are related to the objects (human, vehicles) in the scene.  
\par The workflow of the proposed system is as follows: The input to the proposed model is a UAV aerial video frame. Initially, an object detector is used to detect all the objects in the given video frame.  Further, temporal and appearance features are extracted for every instance of the object of interest (vehicles/humans) detected in the frame. Also, the given input frame is semantically segmented to capture the contextual information around the detected objects. Finally, the contextual features along with the temporal and appearance features are given as input to the inference algorithm. The inference algorithm assigns an anomaly score for every instance of the object of interest (human/vehicle) present in the scene. The object level score is assigned as the frame level anomaly score if a single object of interest is present in the frame. In case of multiple objects present in the frame, the frame-level anomaly score is thus estimated as the maximum of the anomaly score assigned to these objects. 

\subsubsection{Object detection}
\par Normally, anomalous events are associated with the objects in the scene \cite{Ionescu_2019_CVPR}, \cite{doshi2020any}. Therefore, recent work have focused on object of interest in the scene  to identify the presence/absence of anomalous events. Moreover, the detection of an anomalous event linked to an object in the scene allows us to identify the location of the anomalous event in the scene \cite{Ionescu_2019_CVPR}. 

\par In this work, two classes of objects namely vehicles and humans are considered for anomaly detection.  In this study, YoloV3 \cite{redmon2018yolov3} is used to detect these objects in the given scene. YoloV3 is trained from scratch on MUAAD dataset to detect humans and vehicles in each frame of the video.

\subsubsection{Feature extraction}

An anomalous event is generally categorized into three groups namely contextual, temporal and appearance anomaly. Hence, for an accurate prediction of anomalous patterns, it is important to extract contextual, temporal and appearance features. Moreover, in a multi-scene scenario such as videos acquired from a moving UAV, the definition of anomaly depends on the context of the scene. Despite this fact, methods proposed in the literature ignores contextual features. This limits the application of developed methods to multi-scene scenarios. Hence, in the present study, a novel feature extractor is proposed which considers contextual, temporal and appearance features. It is observed that so far these features have not been considered holistically for detecting anomalies in video sequences. However, in the present analysis, for each of the detected object of interest $q$ (vehicle, humans),  the contextual ($f^c_q$), temporal ($f^t_q$) and appearance ($f^a_q$) features are extracted. The final feature descriptor $F_q$  for the object of interest $q$  is the combination of $f^c_q$, $f^t_q$ and $f^a_q$ and is of dimension $22$.
\begin{equation}
    F_q=\{f^c_q,f^t_q,f^a_q\}
\end{equation}The process for extraction of these features for a particular object of interest $q$ is explained below. Note that for the sake for simpler notation, $f^c$, $f^t$ and $f^a$ refers to the contextual, temporal and appearance features respectively for a particular object of interest $q$ unless otherwise specified. 
\begin{figure*}[!ht]
	\begin{tabular}{cccccccc}
		\begin{minipage}{32pt}
			\includegraphics[width=0.8in, height=1in]{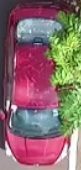}
		\end{minipage}
		&
		\hspace{0.5cm}
		\begin{minipage}{32pt}
			\includegraphics[width=0.8in, 	height=1in]{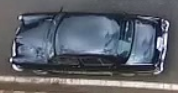}
		\end{minipage}
		&
		\hspace{0.5cm}
		\begin{minipage}{32pt}
			\includegraphics[width=0.8in, 	height=1in]{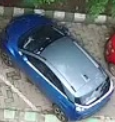}
		\end{minipage}
		&
		\hspace{0.5cm}
		\begin{minipage}{32pt}
			\includegraphics[width=0.8in, 	height=1in]{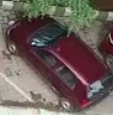}
		\end{minipage}
		&
		\hspace{0.5cm}
		\begin{minipage}{32pt}
			\includegraphics[width=0.8in, 	height=1in]{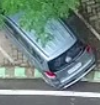}
		\end{minipage}
		&
		\hspace{0.5cm}
		\begin{minipage}{32pt}
			\includegraphics[width=0.8in, 	height=1in]{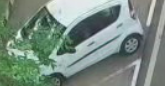}
		\end{minipage}
		&
		\hspace{0.5cm}
		\begin{minipage}{32pt}
			\includegraphics[width=0.8in, 	height=1in]{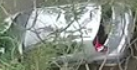}
		\end{minipage}
		&
		\hspace{0.5cm}
		\begin{minipage}{32pt}
			\includegraphics[width=0.8in, 	height=1in]{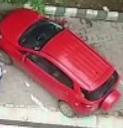}
		\end{minipage}
		\\
		\\
		
		\begin{minipage}{32pt}
			\includegraphics[width=0.8in, height=1in]{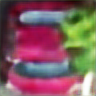}
		\end{minipage}
		&
		\hspace{0.5cm}
		\begin{minipage}{32pt}
			\includegraphics[width=0.8in, 	height=1in]{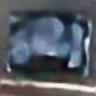}
		\end{minipage}
		&
		\hspace{0.5cm}
		\begin{minipage}{32pt}
			\includegraphics[width=0.8in, 	height=1in]{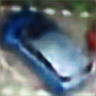}
		\end{minipage}
		&
		\hspace{0.5cm}
		\begin{minipage}{32pt}
			\includegraphics[width=0.8in, 	height=1in]{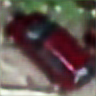}
		\end{minipage}
		&
		\hspace{0.5cm}
		\begin{minipage}{32pt}
			\includegraphics[width=0.8in, 	height=1in]{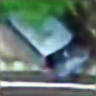}
		\end{minipage}
		&
		\hspace{0.5cm}
		\begin{minipage}{32pt}
			\includegraphics[width=0.8in, 	height=1in]{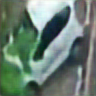}
		\end{minipage}
		&
		\hspace{0.5cm}
		\begin{minipage}{32pt}
			\includegraphics[width=0.8in, 	height=1in]{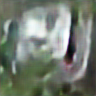}
		\end{minipage}
		&
		\hspace{0.5cm}
		\begin{minipage}{32pt}
			\includegraphics[width=0.8in, 	height=1in]{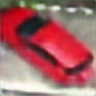}
		\end{minipage}
		\\
		\\
		
	\end{tabular}
	\caption{Appearance Feature Extraction: A few sample input images (top row) and its corresponding reconstructed images (bottom row). Note that a large reconstruction error is observed in few images representing abnormal event. }
	\label{recon}
\end{figure*}

\begin{figure}[!t]
	\centering
	\includegraphics[width=2.8in,height=4in]{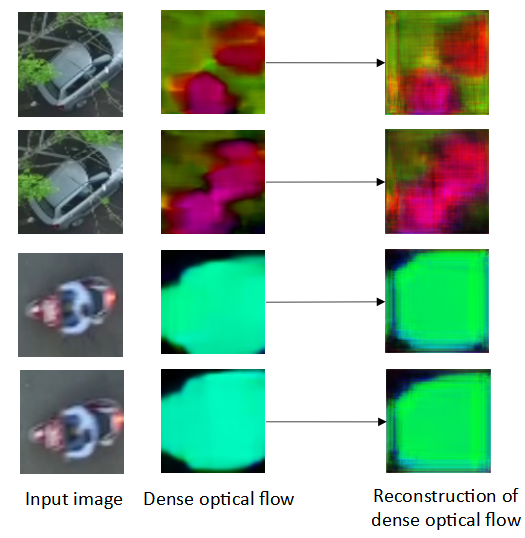}
	
	\caption{Temporal Feature Extraction: Reconstruction of dense optical flow for temporal feature extraction. Note that a large reconstruction error is observed for objects associated with anomalous event.}  
	\label{temp}
\end{figure}
\par \textbf{Temporal features ($f^t$): }The proposed temporal feature extractor relies on the auto-encoder based reconstruction error computed from the optical flow. 

\begin{figure*}[t]
	\centering
	\includegraphics[width=5.8in,height=1.8in]{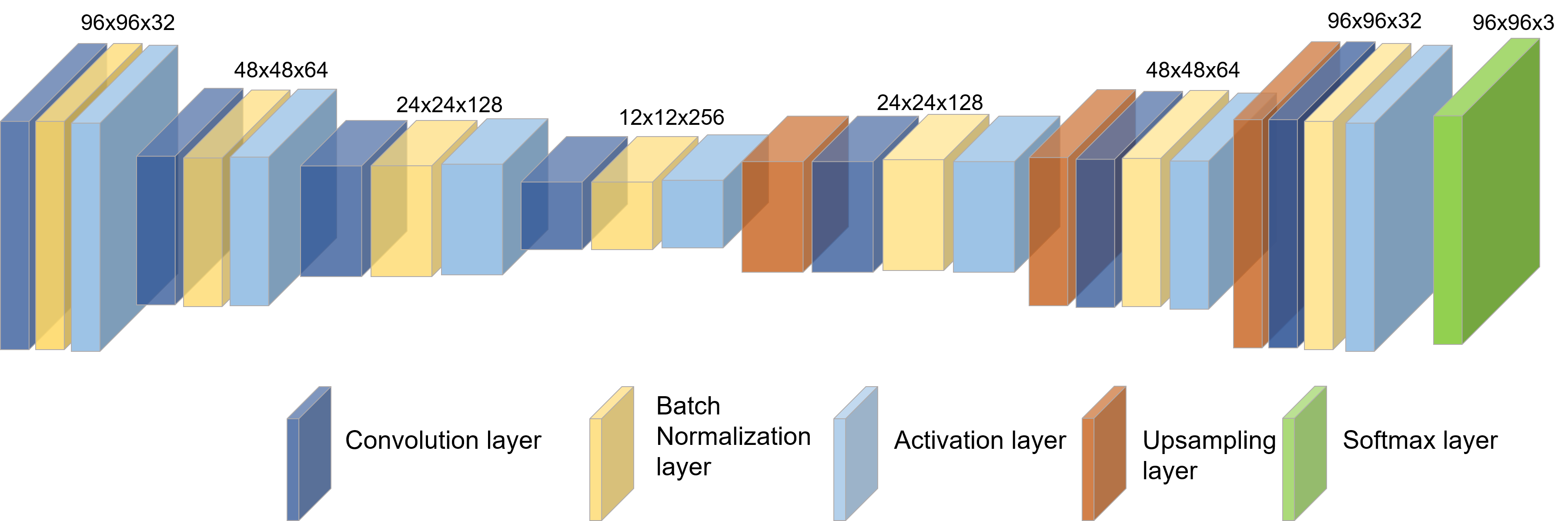}
	\caption{Architecture of the auto-encoder.}
	\label{autoencoder}
\end{figure*}

Initially, the dense optical flow \cite{farneback2003two} of the bounding box corresponding to the object of interest is estimated from the two consecutive image frames. These dense motion vectors are represented as RGB color image patches. Further, these RGB image patches are given as input to the auto-encoders.  The auto-encoders converts the input image to latent space and back to the image domain using an unsupervised learning method. A few sample images of dense optical flow and the corresponding reconstructed flow is shown in Figure \ref{temp}. Here, the auto-encoders are trained on normal motion patterns. Hence, these auto-encoders fails to reconstruct those motion patterns which deviates from normal patterns thus resulting in a higher reconstruction error. The reconstruction error is computed as the absolute difference between the original input RGB image and the reconstructed image (output of auto-encoder) for each individual color channel ($E^{t}_r, E^{t}_g, E^{t}_b$). The proposed auto-encoder for temporal feature extraction has 4 encoding layers and 3 decoding layers. The overview of the proposed auto-encoder is shown in Figure \ref{autoencoder}. Each layer has a convolution layer followed by a batch normalization layer and an activation layer. In the present study, a filter size of 3x3 is utilized. ReLU activation function is used. In the end, softmax layer is applied to reconstruct the input image.   Further,  first-order statistical features such as mean ($S^{t}_1$), variance ($S^{t}_2$), kurtosis ($S^{t}_3$), energy ($S^{t}_4$), skewness ($S^{t}_5$) and entropy ($S^{t}_6$) \cite{sainju2014automated}of the reconstructed image patch along with the reconstruction error ($E^{t}_r, E^{t}_g, E^{t}_b$) produced by the model are considered as the temporal feature vector for the object and is given by: 
\begin{equation}
    f^t=\{E^{t}_r,E^{t}_g,E^{t}_b,S^{t}_1,S^{t}_2,S^{t}_3,S^{t}_4,S^{t}_5,S^{t}_6\}
\end{equation}

\par \textbf{Appearance features ($f^a$): } An appearance feature extractor is defined similar to the temporal feature extractor. In particular, the appearance feature extractor consists of an auto-encoder whose input is the image patch corresponding to object of interest. The model is trained to reconstruct the normal objects. Hence, any new/anomalous objects will produce a higher reconstruction error. A few samples of detected objects and their reconstructed image is shown in Figure \ref{recon}. The reconstruction error in the RGB ($E^{a}_r, E^{a}_g, E^{a}_b$) color plane along with first-order statistical features mean ($S^{a}_1$), variance ($S^{a}_2$), kurtosis ($S^{a}_3$), energy ($S^{a}_4$), skewness ($S^{a}_5$) and entropy ($S^{a}_6$) is considered as appearance feature vector which is given as follows:
\begin{equation}
    f^a=\{E^{a}_r,E^{a}_g,E^{a}_b,S^{a}_1,S^{a}_2,S^{a}_3,S^{a}_4,S^{a}_5,S^{a}_6\}
\end{equation}

\par \textbf{Contextual features ($f^c$): } Semantic segmentation algorithms assigns pixel level labels to each individual pixels in the image and are widely used to realize the context of the scene \cite{cordts2016cityscapes}, \cite{girisha2021uvid}. A similar approach is considered to extract contextual information required for anomaly detection. UVid-Net proposed in \cite{girisha2021uvid}, is used to semantically segment the given UAV aerial video frame. UVid-Net is the state-of-the-art algorithm for semantic segmentation of UAV aerial videos. A brief overview of UVid-Net is provided below: UVid-Net is an encoder-decoder based architecture which incorporates temporal information for semantic segmentation for aerial videos. The two consecutive keyframes as the input to the encoder ensures that the segmentation output contains temporally consistent labels without the need for any additional sequential module. Besides, a modified decoder module consisting of multiplication operation instead of concatenation produces a much finer segmentation result. More details about this architecture can be found in \cite{girisha2021uvid}.

In this work, UVid-Net is trained to segment the given frame into four classes namely greenery, construction, roads and water bodies. These four broad semantic classes help in modeling the background scene information which can be utilized to learn context of the scene. 

\par
\begin{figure}[t]
	\centering
	\includegraphics[width=3.5in,height=1.5in]{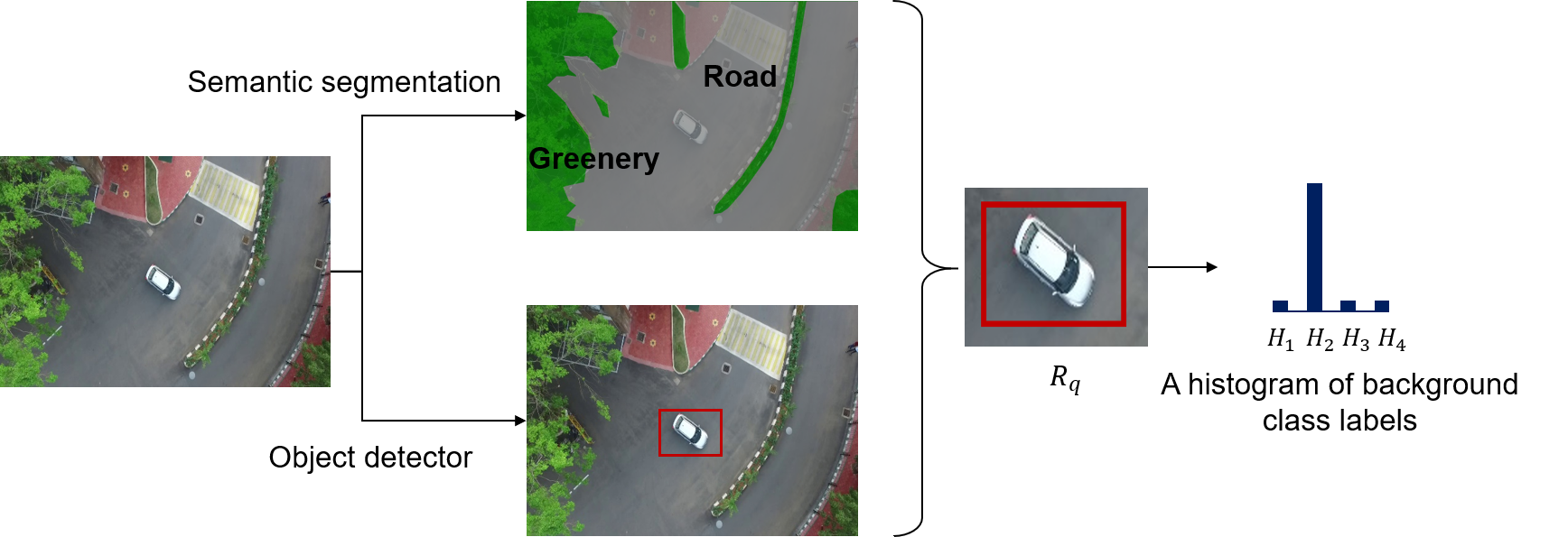}
	\caption{Contextual Feature Extraction: The distribution  of pixels belonging to four background classes is utilized for representing contextual information around the object of interest.  }
	\label{contextual}
\end{figure}

 Generally, the bounding box produced by the object detectors are small and tight around the object. Also, context of the object is determined by its surrounding pixels. Hence, in this work, a small region $R_{q}$ (4 pixel width) around and within the bounding box of the detected object is considered for extracting contextual information. Further, the total number of occurrence of the four class labels (greenery, construction, roads and water bodies) is computed in the identified small region $R_{q}$. This histogram of class labels defines the context around the object and is considered as the contextual feature.  For instance, in case of  object situated on the road, pixels belonging to the road class will have a maximum count in the histogram (Figure \ref{contextual}). Let us represent the four bins of the histogram as $H_{1}, H_{2}, H_{3}, H_{4}$ corresponding to the total number of the occurrence of the pixels belonging to greenery, road, construction and water classes respectively in the small region $R_{q}$. The contextual feature vector is then defined as: 
\begin{equation}
    f^c=\{H_1,H_2,H_3,H_4\}
\end{equation}


\subsubsection{Model Training}
This section describes the training procedure adopted for training the object detector, auto-encoders and UVid-Net for object detection and feature extraction. It may be noted that object detector, temporal, appearance and contextual feature extractor modules are trained independently of each other as described below. In the present study, two generic classes of objects (vehicles, humans) are considered for detecting anomalies. 

\par \textbf{Object detector:}The YoloV3 \cite{redmon2018yolov3} model is trained on UAV aerial images to detect vehicles and humans. To this end, 300  UAV aerial images are selected from MUAAD dataset and annotations are provided for objects vehicles and humans.  These images and annotations are further utilized for training the YoloV3 model. The K-Means algorithm is used to determine the sizes of 9 anchor boxes. The model is trained for 250 epochs with Adam optimizer and learning rate set to 0.01. 
\par  \textbf{Feature extractor:} From the training data, objects representing normal patterns are manually cropped to create a training set for appearance auto-encoder. In addition, dense optical flow for these cropped images is estimated to create a training set for temporal auto-encoders.  The two auto-encoders are trained separately using Binary cross-entropy loss function and Adam optimizer. Also, data augmentation is employed to increase the training dataset size and prevent the model from overfitting. Various image transformations such as flipping, rotation, translation and shearing are utilized to augment data.  Subsequently, these augmented data is utilized for training the auto-encoders.  For contextual feature extractor, the UVid-Net model is trained on ManipalUAVid dataset \cite{girisha2021uvid} to semantically segment the given aerial image into four classes. The categorical-cross-entropy loss is used with Adam optimizer to train the UVid-Net.

\subsubsection{Inference}

\par In the present study, a new inference algorithm is proposed for detecting anomalies. In the literature, it is observed that the anomaly detection models are solely trained on normal patterns due to the rarity of abnormal events. However, these algorithms may fail to detect local anomalies since the decision boundaries are determined based on normal patterns exclusively. The local anomaly is referred to as the anomalous events which closely resembles a normal event. Hence, in the present study, a  learning strategy is proposed to address the issue of local anomalies. In this strategy, the inference algorithm is trained on a larger set of normal patterns and a limited set of anomalous patterns.  The inclusion of a smaller set of anomalous patterns allows the model to identify better decision boundaries and aids in improved accuracy. 

\par The proposed method is inspired by the inference algorithm presented in \cite{Ionescu_2019_CVPR}. In \cite{Ionescu_2019_CVPR}, the authors proposed to train $K$ SVM classifiers on $K$ clusters of normal patterns to create a model for $K$ normal events. Subsequently, the score from binary classifiers in one-vs-rest scheme is utilized to determine the anomalous events. The use of one-vs-rest scheme creates artificial dummy abnormal events (one normal vs others dummy abnormal events). However, this approach considers only normal patterns to identify decision boundaries. In this work, we propose incorporating a limited set of anomalous event in multi-class classification for anomalous activity detection to learn better discriminative decision boundaries. This work first train two disjoint sets of SVM classifiers, one trained only on normal events and other trained only on abnormal events. During inference, the maximum classification scores from these two sets of classifiers for the test sample are compared to identify the anomalous events.

\par The $22$ dimension feature vector extracted by the proposed feature extractor is used to train the inference algorithm. Let the given training set $T$ consists of $M$ normal samples and $N$ anomalous samples, where $N<<M$. Further, the K-Means algorithm is employed to cluster the $M$ normal samples and $N$ anomalous samples into $K^1$ and $K^2$ groups respectively. Subsequently, we initialize an SVM classifier for each of the  $K^1+K^2$ clusters which results in $K=K^1+K^2$ SVM classifiers. The SVM classifier initialized for normal patterns is trained to classify normal patterns while SVM classifiers trained on anomalous patterns are trained to classify anomalous patterns. While training $i^{th}$ SVM classifier, the samples belonging to other clusters $K-1$ clusters are considered anomalous patterns. Hence, each SVM classifier is trained as a binary classifier. Since these classifiers are trained on the training set $T$ which consists of both normal and anomalous patterns, the decision boundaries identified will be tuned to separate local anomalies. Note that the sample here refers to the feature vector corresponding to a particular object. 

During the test time, for every detected object in the given frame, we extract the feature vector $F$. This feature vector is classified by $K$ SVM classifiers. Further, we calculate the scores as follows:  
\begin{eqnarray}
	\alpha=max(m_1,m_2,...m_{k^1})\\
	\beta  = max(n_1,n_2,...n_{k^2})
\end{eqnarray} 
here, $\alpha$ represents the maximum score of $K^1$ SVM classifiers of normal patterns, while $\beta$ represents the maximum score of $K^2$ SVM classifiers of anomalous patterns. A value of $\alpha$ significantly higher than $\beta$ indicates that the object represents an normal event, while $\alpha < \beta$ indicates an abnormal event corresponding to the object. Therefore, the samples satisfying the following criterion is regarded as normal objects: 

\begin{equation}
\label{Eqn:CondNorm}
    \alpha > \beta ~~AND ~~\alpha > \mu
\end{equation}
where $\mu$ is a parameter determined experimentally and $AND$ is a logical AND operator. Similarly, the samples satisfying the following condition is considered as abnormal objects: 
\begin{equation}
\label{Eqn:CondAbNorm}
    \alpha < \beta~~ AND~~ \beta > \nu
\end{equation}
where $\nu$ is a parameter determined experimentally and $AND$ is a logical AND operator. Moreover, if any samples does not satisfy any of the above two conditions (Eqn \ref{Eqn:CondNorm} and \ref{Eqn:CondAbNorm} ), it is regarded as anomalous object. The above conditions ensure that the object which has been classified by one of the $K^1$ classifiers with a high confidence (high $\alpha$) is regarded as an object associated with normal event. Similarly, the object for an abnormal event would be classified by one of the $K^2$ classifiers with high confidence (high $\beta$). Moreover, it is not possible to model all the possible anomalous activities. Therefore, an object classified as anomalous with less confidence (low $\beta$) is also considered as anomalous event. 

The final frame level anomaly score is computed as the maximum of all the score obtained by the object present in the frame. 

\par In general, a new test sample may belong to one of the three groups namely \emph{normal pattern},  \emph{anomalous pattern} and \emph{unknown pattern}. Unknown patterns are those patterns that are new to the model and may be anomalous. Hence, these patterns should also raise an alarm. Moreover, the sensitivity of anomaly detectors to normal and anomalous patterns plays an important role in providing security. In this regard, this paper presents a simple logic to control the sensitivity of the model to normal and anomalous patterns. Here, the threshold value $\mu$ and $\eta$ are used to regulate the model sensitivity to normal and anomalous patterns respectively.  A given test sample is considered as a normal pattern if $\alpha > \beta$ and $\alpha>\mu$.  Further, the given test sample is considered anomalous if $\alpha < \beta$ and $\beta>\eta$ . Else, the test sample is considered as an unknown pattern. By increasing the value of $\mu$ and $\eta$, the model's responsiveness to the normal and abnormal patterns reduces. Hence, these parameters can be used to control the effectiveness of the anomaly detector depending on the application. The selection of these parameters are discussed in section \ref{parameters}.

\section{Results and discussion}
\label{res}

The proposed algorithm is evaluated on the MUAAD dataset containing the videos acquired from a moving UAV at multiple locations. As discussed earlier, the existing datasets (Ped1, Ped2, ShanghaiTech) focuses on single scene videos acquired from static cameras with little or no camera motion. Since, the proposed algorithm is designed for a multi-scene scenario, it wont be prudent to evaluate the performance of the proposed algorithm on existing single scene datasets. Moreover, the existing datasets does not provide pixel level labels required for extracting contextual information.

This section first presents the performance metrics (Section \ref{SubSec:metrics}), and the approach utilized to select the parameters of the proposed approach (Section \ref{parameters}). Subsequently, the performance of the proposed contextual feature extractor is studied in Section \ref{SubSec:EvaluationOfContextual}, while the performance of the proposed inference algorithm is discussed in Section \ref{SubSec:EvaluationOfInference}. Finally, the proposed approach is compared with the existing methods in Section \ref{SubSec:ComparisonWithOtherMethods}.   

\begin{figure*}[t]
	\begin{tabular}{cc}
		\begin{minipage}{32pt}
			\includegraphics[width=3in, 	height=1.8in]{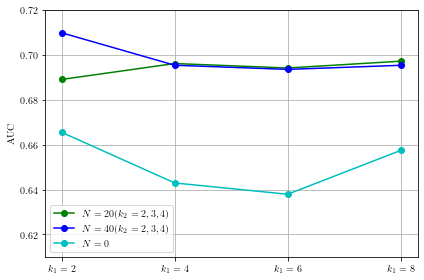}
			\centering{(a)}
		\end{minipage}
		&
		\hspace{7cm}
		\begin{minipage}{32pt}
			\includegraphics[width=3in, 	height=1.8in]{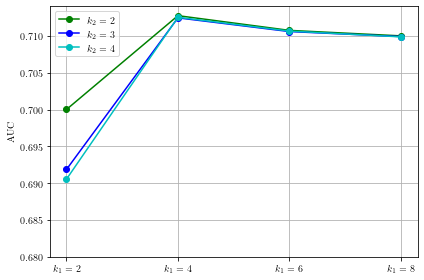}
			\centering{(b)}
		\end{minipage}
		\\
		\begin{minipage}{32pt}
			\includegraphics[width=3in, 	height=1.8in]{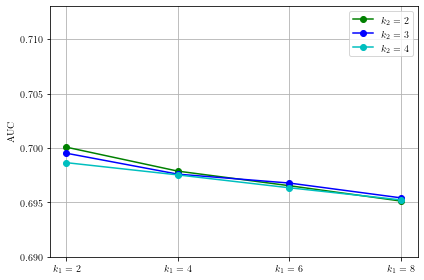}
			\centering{(c)}
		\end{minipage}
		&
			\hspace{7cm}
		\begin{minipage}{32pt}
			\includegraphics[width=3in, 	height=1.8in]{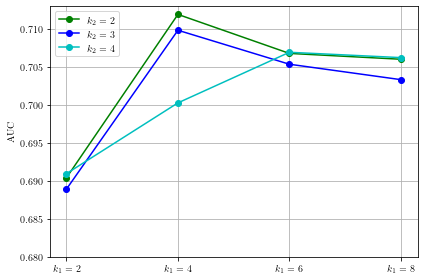}
			\centering{(d)}
		\end{minipage}
		\\
		\\
	\end{tabular}
	\caption{AUC comparison of proposed method with different values of $N$, $K_1$ and $K_2$. (a) shows the AUC comparison with $N=\{0,20,40\}$. (b), (c) and (d) shows the AUC comparison at $N=60$, $N=80$ and  $N=100$ respectively. }
	\label{NK1K2}
\end{figure*}

\subsection{Performance metric}
\label{SubSec:metrics}
\par Several recent works of literature used Area under the ROC Curve (AUC) metric to evaluate the performance of anomaly detector  \cite{Ionescu_2019_CVPR},\cite{doshi2020any},\cite{cong2011sparse},\cite{liu2018classifier}. Following their procedure, in the present study, the AUC metric is used for evaluating the performance of the proposed method. In this method, the frame-level anomaly score is compared against the ground truth frame level annotation, to compute the AUC. Specifically, the frame level AUC is computed from the frame level total positive rate and false positive rate. As discussed earlier, a frame is considered anomalous if it contains at least one object with abnormal activity.

\subsection{Parameter selection}
\label{parameters}
\par This section discusses the selection of parameters used in the proposed model such as the number of normal  $K^1$ and abnormal   $K^2$  clusters and the thresholds $\mu$ and $\eta$. It may be noted that a normal/anomalous sample for the inference model refers to the feature vector associated with normal/anomalous object of interest. 
\par \textbf{Selection of $K^1$ and $K^2$:} The parameters $K^1$ and $K^2$ represent the number of clusters in normal and anomalous training sample and has an influence on the performance of the model . The values for these two parameters are determined experimentally.  The AUC metric is studied for different pair of values of $K^1$ and $K^2$. The $K^1$ value is selected from the set $\{2,4,6,8\}$ while $K^2$ value is selected from the set $\{2,3,4\}$and a grid search like method is employed to find the optimal value. It may be noted that a lower values of the number of anomalous cluster $K^2$ was considered due to limited number of anomalous samples present in the dataset.  The maximum AUC of 0.712 was observed for $K^1 = 4$  and $K^2 = 3$ (Figure \ref{NK1K2} (b)).

Given the limited number of anomalous samples ($N$) available in the dataset, we also studied the effect on $K^1$ and $K^2$ due to variation in the number of anomalous samples $N$. Figures \ref{NK1K2} shows the AUC obtained by varying $N$, $K^1$ and $K^2$. It can be observed that a maximum AUC of $0.712$ is obtained for    $N=60$, $K^1=4$ and $K^2=3$. This AUC is slightly higher than $N=100$, $K^1=4$ and $K^2=2$. Also, it is observed that the AUC obtained for $K^2=\{2,3,4\}$ are closer by at $N=60$ and $k^1=4$. Therefore, in the present study, $K^1$ and $K^2$ value is determined to be $4$ and $3$ respectively.

\par \textbf{Selection of $\mu$ and $\eta$:}  The parameters $\mu$ and $\eta$ (Equations \ref{Eqn:CondNorm}, \ref{Eqn:CondAbNorm} ) determines the sensitivity of the proposed inference model. A given pattern is considered normal if the the maximum score of normal SVM classifiers ($\alpha$) is greater than the maximum score of abnormal SVM classifiers ( $\beta$) and the threshold $\mu$. Consequently, regulating the $\mu$ value allows us to control the sensitiveness of the model to normal patterns. For example, a higher value of  $\mu$, determines the given pattern as normal if the confidence score ($\alpha$) is greater than $\mu$.  This allows the model to identify normal patterns with more confidence. On the contrary, reducing the $\mu$ value allows the model to determine normal patterns with a lower degree of confidence. Similarly, threshold $\eta$ controls the model's sensitivity to anomalous patterns. Hence, $\mu$ and $\eta$ values should be selected carefully. 
\par
\begin{figure}[t]
	\centering
	\includegraphics[width=3in,height=1.7in]{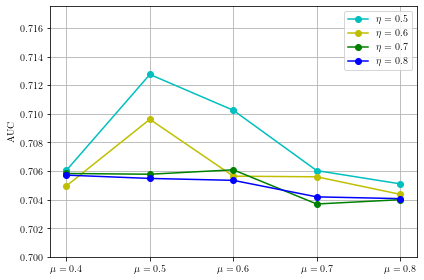}
	
	\caption{AUC comparison of proposed method with different values of $\mu$ and $\eta$.}
	\label{thresh}
\end{figure}

 In the present study, $\mu$ value is selected from set $\{0.4, 0.5, 0.6, 0.7, 0.8\}$.  Concerning anomaly detection, it is always desirable to have a model with high sensitivity to anomalous patterns. Hence, in the present study, $\eta$ is selected from a lower set of threshold values $\{0.5, 0.6, 0.7, 0.8\}$. 
The grid search method is used to determine the optimal value of $\mu$ and $\eta$. This experiment is conducted by setting the values of $N$, $K^1$ and $K^2$ to 60, 4 and 3. The AUC of the proposed model with different values of $\mu$ and $\eta$ is estimated and is plotted in Figure \ref{thresh}. It was found that the AUC of the model reaches $0.0.712$ when $\mu$ is set of $0.5$ and $\eta$ set to $0.5$. Since the model is trained on a smaller set of anomalous patterns, reducing the sensitivity of the model to anomalous patterns (increasing the $\eta$) reduces the AUC of the model. Hence, for a higher value of $\eta$, the AUC of the model reduces. In the present study, $\mu$ and $\eta$ values are determined to be $0.5$ and $0.5$ respectively.

\subsection{Evaluation of contextual knowledge}
\label{SubSec:EvaluationOfContextual}
The performance of anomaly detection algorithms in a multi-scene scenario depends on contextual knowledge about the scene. The proposed work uses appearance features to realize the appearance change of the object while contextual feature builds a summary of the background scene information around the object. To this end, in the present study, a given UAV aerial scene is semantically segmented to realize the layout of the scene. Subsequently, a histogram of class labels is calculated around the object. Finally, this histogram is considered as a feature vector along with the appearance and temporal features to describe the object. This feature vector is further given to the inference algorithm to decide on the anomaly.  In the present study, an experiment is executed to evaluate the influence of contextual knowledge on the performance of anomaly detection. In this experiment, the performance of the proposed model is compared \textit{with} contextual information and \textit{without} contextual information.  Initially, the performance of the model is evaluated using appearance and temporal features. Subsequently, the performance of the model is evaluated using contextual, appearance and temporal features. In this experiment, the proposed inference algorithm is utilized to infer the anomalies. The AUC of the proposed model with and without contextual information is shown in Table \ref{ct}. It was found that there was a significant improvement in AUC (0.712 vs 0.675) of the proposed model with contextual knowledge. This result is significant as it demonstrates the importance of contextual knowledge in a multi-scene scenario. 
\par
\begin{figure}[t]
	\centering
	\includegraphics[width=3in,height=2in]{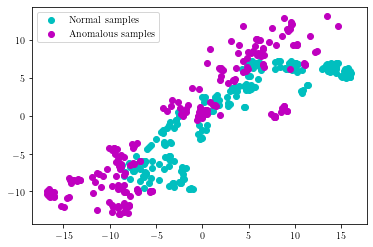}
	
	\caption{Plot of normal and anomalous feature vectors without contextual knowledge. }
	\label{WOSS}
\end{figure}

\begin{figure}[t]
	\centering
	\includegraphics[width=3in,height=2in]{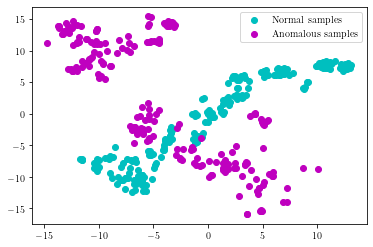}
	
	\caption{Plot of normal and anomalous feature vectors with contextual knowledge. }
	\label{WSS}
\end{figure}

  To qualitatively evaluate the importance of contextual knowledge, we visualized the feature vector with and without contextual knowledge. A total of random  $400$ sample patterns are considered among which $200$ belongs to normal patterns and $200$ belongs to anomalous patterns.   Let the feature vector with contextual knowledge be represented as $F_{1}$ and without contextual knowledge be represented as $F_{2}$.  The dimension of the considered feature vectors $F_{1}$ and $F_{2}$ are $22$ and $18$ respectively. Dimensionality reduction is essential to visualize these feature vectors. To this end, Principal Component Analysis (PCA) is employed to reduce the feature dimensions of $F_{1}$ and $F_{2}$ to two Subsequently, we plotted the features in 2D space. This is shown in Figure \ref{WOSS} and Figure \ref{WSS}. From these figures, it is observed that the feature vector $F_{2}$ discriminates normal and anomalous patterns efficiently. Hence, the inclusion of contextual knowledge aids one-class SVMs to determine better decision boundaries. This further substantiates the improvement in AUC of the model with contextual information. This experiment demonstrates that importance of contextual knowledge in the detection of anomalous patterns from the multi-scene scenario.  
\par
\begin{figure*}[t]
	\centering
	\includegraphics[width=7in,height=3.4in]{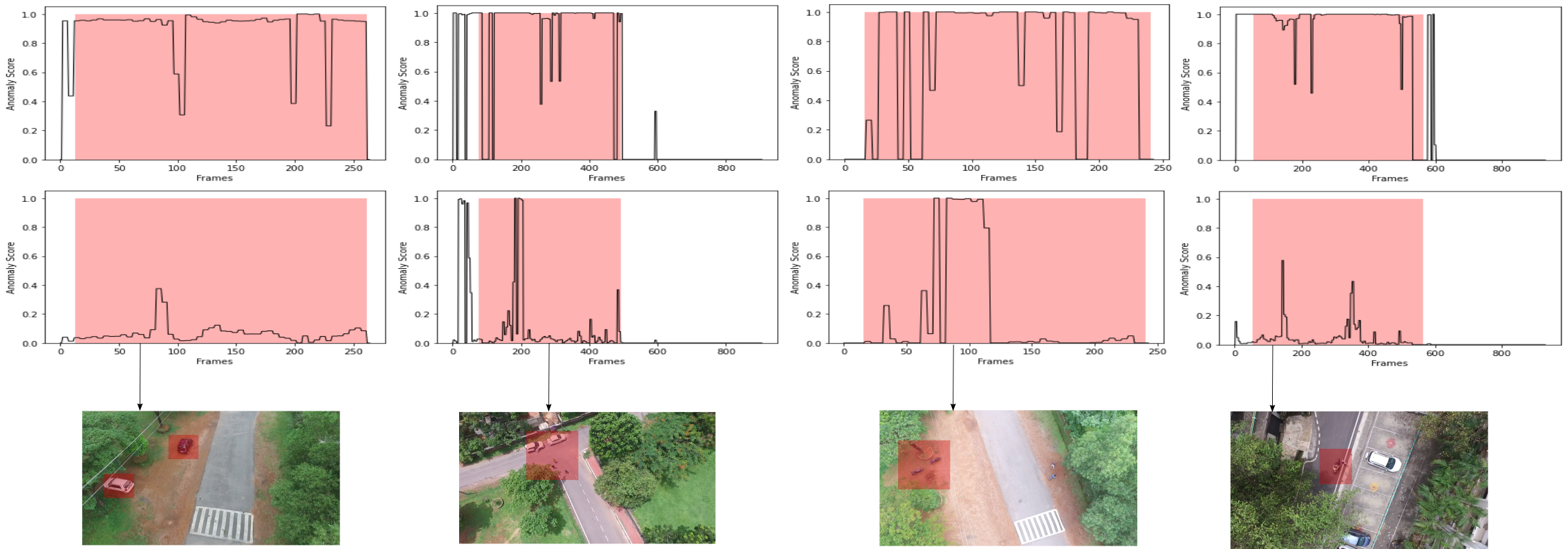}

	\caption{Anomaly score comparison of proposed model with and without contextual knowledge. The red region in the plot indicates the anomalous frames. The first row show the frame-level anomaly score of the proposed model with contextual knowledge. Second row shows the frame-level anomaly score of the proposed model without contextual knowledge. The bottom row shows a sample frame from the video with the anomalous object. }
	\label{anomaly_score}
\end{figure*}

 Further, we compared the anomaly score at the frame level of the proposed model with and without contextual knowledge. Figure \ref{anomaly_score}, shows the frame-level AUC comparison of the proposed model on few multi-scene aerial video samples. These video samples constitute contextual anomalies such as stationary vehicles on the greenery, road.  It is observed that the proposed model with contextual knowledge produces a higher anomaly score for anomalous frames. However, the proposed model without contextual knowledge produces a lower anomaly score.  This result is anticipated since appearance and temporal features alone are not adequate enough to detect contextual anomalies. However, the proposed feature extractor which encapsulates contextual, appearance and temporal features holistically, can effectively detect these anomalies. Besides, the inclusion of contextual information helps in better identification of anomalous frames as compared to the model without any contextual information. It can be observed in Figure \ref{anomaly_score} that a higher number of anomalous frame are correctly classified as anomalous by incorporating contextual information.   This experiment demonstrates the importance of contextual knowledge in multi-scene anomaly detection.

\begin{table}[t]

\centering	
	\caption{AUC comparison of proposed method with and without contextual knowledge.}
	\label{ct}
\begin{tabular}{|c||c|}
	\hline
	
	\textbf{Proposed method} & \textbf{AUC}\\
	\hline
	\hline
	With contextual knowledge  &\textbf{0.712}\\
	\hline
	Without contextual knowledge& 0.6753\\
	\hline

\end{tabular}
\end{table}

\subsection{Evaluation of inference algorithm}
\label{SubSec:EvaluationOfInference}
\begin{figure*}[!ht]
	\begin{tabular}{cc}
		\begin{minipage}{32pt}
			\includegraphics[width=3in, 	height=1.8in]{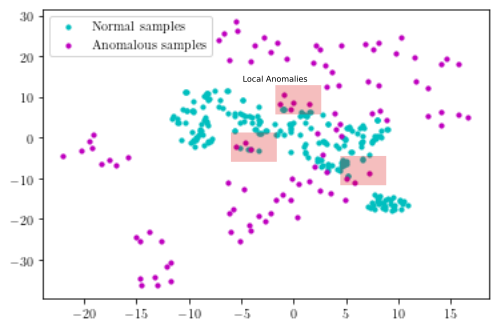}
			\centering{(a)}
		\end{minipage}
		&
		\hspace{7cm}
		\begin{minipage}{32pt}
			\includegraphics[width=3in, 	height=1.8in]{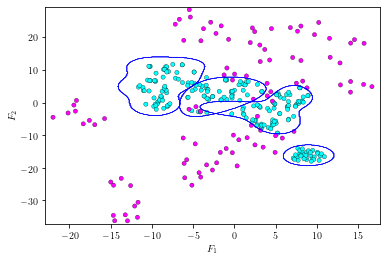}
			\centering{(b)}
		\end{minipage}
		\\
		\begin{minipage}{32pt}
			\includegraphics[width=3in, 	height=1.8in]{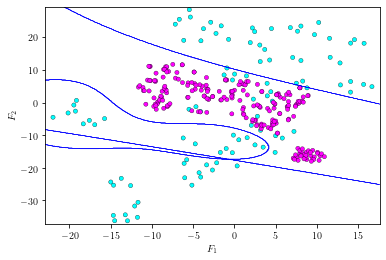}
			\centering{(c)}
		\end{minipage}
		&
		\hspace{7cm}
		\begin{minipage}{32pt}
			\includegraphics[width=3in, 	height=1.8in]{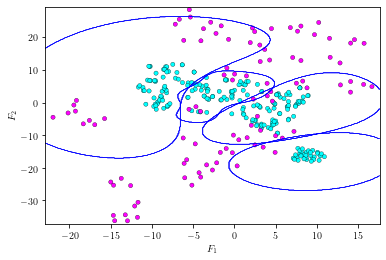}
			\centering{(d)}
		\end{minipage}
		\\

		\\
	\end{tabular}
	\caption{Visualization of decision boundaries of SVM classifiers. (a) shows the scatter plot of normal and anomalous samples. (b) shows the decision boundaries of SVM classifiers trained on normal and few samples of anomalous patterns. (c) shows the decision boundaries of SVM classifiers trained on anomalous samples. (d) show the decision boundaries of SVM classifiers trained on normal samples only.}
	\label{decision_boundaries}
\end{figure*}

A majority of existing woks utilizes only normal pattern to model the normal events. Hence, the decision boundaries are tuned to identify normal patterns and may fail to detect local anomalies. For example, in the case of UAV aerial videos, a car parked in the parking area and on the road may have similar appearance characteristics. However, a car parked on road is an example of local anomalies since its attributes are closer to normal patterns. Hence, an inference algorithm trained only on normal patterns may not be sufficient to distinguish these local anomalies. In the present study, we argue that training the model on few samples of known anomalous patterns can significantly improve the performance of the model.  In this context, we visualized the decision boundaries of $K^1$ and $K^2$ SVM classifiers. To this end, we considered $300$ normal samples and $100$ anomalous samples. Figure \ref{decision_boundaries} (a) shows the scatter plot of considered samples. Few examples of local anomalies are highlighted in RED. An inference algorithm trained only on normal patterns may not distinguish these samples. However, we can observe that the decision boundaries identified by $K^1$ SVM classifiers are effectively separating local anomalies since they are trained using both normal and few anomalous samples. Hence, the proposed model can classify local anomalies more accurately. This result validates the assumption of considering few anomalous samples in identifying the decision boundaries. 

\subsubsection{Evaluating the influence of number of anomalous events} 
\begin{figure}[t]
	\centering
	\includegraphics[width=3in,height=1.7in]{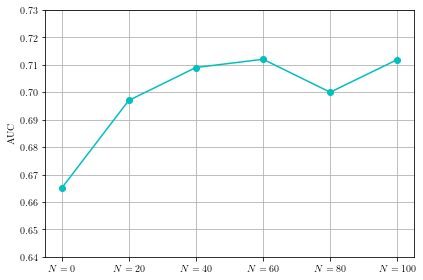}
	
	\caption{AUC comparison of proposed model with different values of N. The best $K_1$ and $K_2$ values are considered for the comparison of the performance.}
	\label{fig_N}
\end{figure}

The rarity of anomalous samples poses a significant challenge for the development of anomaly detection algorithms. The proposed work utilizes few samples of anomalous patterns along with normal patterns to determine the decision boundaries.  The inclusion of few anomalous samples allows the SVM classifiers to determine accurate decision boundaries thereby improving anomaly detection.  In this context, an experiment is performed to learn the influence of anomalous samples on anomaly detection. In this experiment, the proposed inference algorithm is evaluated with different values of number of anomalous events ($N$). The N value is selected from the set $\{0, 20, 40, 60, 80, 100\}$. The M value (number of normal training samples) is set to $300$. In every iteration of the experiment, $N$ anomalous training samples are selected randomly. The grid search approach is utilized to determine the parameters of SVM classifiers. Grid search is a hyperparameter tuning method that implements exhaustive searching to determine the hyperparameters. Figure \ref{fig_N}, shows the AUC of the proposed method with different values of $N$.  At $N=0$, the model is trained only on normal samples. Hence, the decision boundaries may fail to accurately classify anomalous samples. Consequently, it is seen that the proposed method achieves an AUC of $0.665$ at $N=0$. However, as the N value increases, the AUC of the algorithm increases and reaches an AUC of $0.712$ at $N=60$.  An important observation here is that at $N=20$, the model achieved an AUC of $0.697$ which is notably greater than the AUC ($0.665$) obtained at $N=0$. This is a significant result since it demonstrates the importance of a \emph{few anomalous samples} in identifying the decision boundaries for accurate anomaly detection. 

\par As discussed earlier, the proposed inference algorithm is inspired by the work of \cite {Ionescu_2019_CVPR} which formulated abnormal event detection as multi-class classification problem. Therefore, we also evaluate the proposed inference algorithm by comparing it with the inference algorithm proposed in \cite{Ionescu_2019_CVPR}. In this experiment, we utilized the proposed feature extractor to extract contextual, temporal and appearance features. Subsequently, the performance of proposed and existing \cite{Ionescu_2019_CVPR}  inference algorithms are studied for inferring the anomaly. Note that the existing inference algorithm is only trained on normal patterns with the number of clusters set to $5$. The proposed inference algorithm is trained on normal and known anomalous patterns with $K^1$ set to $4$ and $K^2$ set to $3$. Table \ref{it} shows the AUC results of the two inference algorithm on the proposed UAV anomaly detection dataset with different feature extractors. It is found that the temporal features produced higher AUC as compared to contextual and appearance features on the existing and proposed inference algorithms. This result is justified by the fact that the MUAAD dataset has higher temporal anomalies due to object motion. Further, it is seen that contextual features alone are not sufficient to detect anomalies in UAV aerial videos. However,  by considering the appearance, temporal and contextual features holistically, the performance of the existing and the proposed inference algorithm improved significantly. Specifically, the proposed and the baseline inference algorithm achieved an AUC of 0.712 and 0.648 respectively. It was found that the proposed inference algorithm produced a significant improvement in AUC ($7\%$) as compared to \cite{Ionescu_2019_CVPR}. This finding confirms that the decision boundaries identified by the proposed inference algorithm are tuned to separate both normal and anomalous patterns.  
Furthermore, the proposed feature extractor has improved the performance of both proposed and baseline inference algorithm. This improvement is consistent and hence validates the proposed feature extractor for UAV anomaly detection.

\begin{table}[!t]
	
	\centering	
	\caption{AUC comparison of proposed model with different inference algorithm feature extractor.}
	\label{it}
	\begin{tabular}{|c||c|}
		\hline
		
		\textbf{Inference method} & \textbf{AUC}\\
		\hline
		\hline
		Ionescu et al.(Appearance features)  &0.587\\
		\hline
		Ionescu et al.(Temporal features)  &0.6028\\
		\hline
		Ionescu et al.(Contextual features)  &0.566\\
		\hline
		Ionescu et al.(Proposed feature extractor)  &0.648\\
		\hline
		Ours (Appearance features)  &0.5973 \\
		\hline
		Ours (Temporal features)  &0.6332 \\
		\hline
		Ours (Contextual features)  &0.5513 \\
		\hline
		Ours (Proposed feature extractor)   &\textbf{0.712} \\
		\hline
		
	\end{tabular}
\end{table}

\subsection{Comparative study} 
\label{SubSec:ComparisonWithOtherMethods}
\par In the present study, two semi-supervised anomaly detection algorithms are considered as baseline models. In the first method \cite{Ionescu_2019_CVPR}, the authors proposed to use auto-encoders to extract features. Subsequently, SVM based inference algorithm is utilized to infer the anomalies. In \cite{doshi2020any}, the authors proposed a continual learning approach using the KNN algorithm to detect anomalies in videos. In addition, the proposed method is also compared with the approaches proposed in \cite{astrid2021synthetic} and \cite{li2021two}. Both these methods are based on auto-encoders and are non-object centric. All these methods are designed for CCTV videos. Here it is important to note that, the feature extractor of the proposed model is improved upon the feature extractor proposed in \cite{Ionescu_2019_CVPR} and \cite{doshi2020any}. Furthermore, the inference algorithm is designed based on the idea proposed in \cite{Ionescu_2019_CVPR}. 
\par The AUC results of the proposed and compared methods on UAV anomaly detection dataset is given in Table \ref{compt}. It was found that the methods proposed in \cite{Ionescu_2019_CVPR} and \cite{doshi2020any} achieved an AUC of $0.568$ and $0.641$ respectively on UAV anomaly detection dataset. The methods proposed in \cite{astrid2021synthetic} and \cite{li2021two} achieved an AUC of $0.487$ and $0.359$  respectively on MUAAD dataset.  These methods \cite{astrid2021synthetic}, \cite{li2021two}  are designed for CCTV videos where the background is constant with little or no camera motion. However, UAV aerial videos contains significant camera motion. Hence, the temporal features extracted by these methods may produce false positives in the presence of camera motion. In addition, the methods proposed in \cite{astrid2021synthetic} and \cite{li2021two} are non-object centric. Hence, a higher reconstruction error will be observed in a multi-scene scenario such as UAV aerial video. These high reconstruction error would be due to variation in background information and need not represent an actual anomalous event.  This contributes in higher false positives and lower AUC. This result is significant as it highlights the effectiveness of object centric methods in a multi-scene scenario.  The proposed \textbf{object-centric method} uses auto-encoders and optical flow to extract temporal features. Since the auto-encoder is trained on normal motion patterns, it produces in higher reconstruction error for anomalous patterns. Also, the proposed method calculates temporal features concerning the object bounding box which further reduces the influence of camera motion on temporal features. It can be observe that the proposed method achieved an AUC of $0.712$ which is significantly greater than baseline models. 

\par Furthermore, the UAV anomaly detection dataset has videos taken from various locations (multi-scene scenarios). Also, each video constitutes camera motion. Hence, the context of the scene differs within a video. In these situations, realizing the context of the scene aids in improving the performance of anomaly detection. Despite this interest, the baseline models ignore contextual features. However, the proposed method captures the contextual knowledge required for context-aware anomaly detection. This result is significant since the definition of anomaly is dependent on the context of the scene which is generally ignored in the literature. The proposed method highlights the importance of contextual information required for the accurate detection of anomalies in multi-scene scenarios such as UAV.
\begin{table}[!t]

\caption{AUC comparison of proposed and other methods.}
\label{compt}
\centering

\begin{tabular}{|c||c|}
\hline

\textbf{Algorithm} & \textbf{AUC}\\
\hline
\hline
Ionescu et al.\cite{Ionescu_2019_CVPR}  & 0.568\\
\hline
Doshi et al. \cite{doshi2020any}& 0.641 \\ 
\hline
Li T et al. \cite{li2021two}&0.359\\
\hline
Astrid et al.\cite{astrid2021synthetic}& 0.487\\
\hline
Ours & \textbf{0.712} \\
\hline

\end{tabular}
\end{table}

\section{Conclusion}
\label{con}
This work is intended to develop a video anomaly detection algorithm for UAV aerial videos. A new multi-scene UAV anomaly detection dataset is proposed to address the lack of standard datasets for anomaly detection in UAV surveillance videos. The proposed dataset has scene variations and camera motion that provides a standard challenging platform for researchers to develop and evaluate their algorithms. Frame-level annotations are provided for 52 UAV aerial videos. Further, two baseline models are implemented and evaluated for validating the dataset.

\par This study also introduces a new UAV video anomaly detection algorithm that holistically uses contextual, temporal and appearance features to detect the anomalies. Furthermore, a novel inference algorithm is presented that uses a few-shot learning strategy to infer the anomalies. The proposed algorithm achieved an AUC of 0.712 which is significantly greater than the compared baseline models. The extensive evaluation of the proposed model revealed that contextual knowledge plays a significant role in multi-scene video anomaly detection. The inclusion of contextual knowledge leverages the performance of video anomaly detection algorithms. Furthermore, this study demonstrated that the incorporation of few known anomalous samples in the training process can prominently improve the performance of anomaly detection. This study highlights the importance of contextual knowledge and learning strategy for video anomaly detection.

\section{Acknowledgement}
\par This research was supported by Manipal Academy of Higher Education for the dataset generation inside the campus using drones (MUAAD, ManipalUAVid). A provisional Indian patent has been filed for the developed system. Patent application number: 202241010616. Date of filing: 28/02/2022.

\bibliographystyle{IEEEtran}
\bibliography{reference}

\begin{thebibliography}{10}
\providecommand{\url}[1]{#1}
\csname url@samestyle\endcsname
\providecommand{\newblock}{\relax}
\providecommand{\bibinfo}[2]{#2}
\providecommand{\BIBentrySTDinterwordspacing}{\spaceskip=0pt\relax}
\providecommand{\BIBentryALTinterwordstretchfactor}{4}
\providecommand{\BIBentryALTinterwordspacing}{\spaceskip=\fontdimen2\font plus
\BIBentryALTinterwordstretchfactor\fontdimen3\font minus
  \fontdimen4\font\relax}
\providecommand{\BIBforeignlanguage}[2]{{%
\expandafter\ifx\csname l@#1\endcsname\relax
\typeout{** WARNING: IEEEtran.bst: No hyphenation pattern has been}%
\typeout{** loaded for the language `#1'. Using the pattern for}%
\typeout{** the default language instead.}%
\else
\language=\csname l@#1\endcsname
\fi
#2}}
\providecommand{\BIBdecl}{\relax}
\BIBdecl

\bibitem{ramachandra2020survey}
B.~Ramachandra, M.~Jones, and R.~R. Vatsavai, ``A survey of single-scene video
  anomaly detection,'' \emph{IEEE Transactions on Pattern Analysis and Machine
  Intelligence}, 2020.

\bibitem{chalapathy2019deep}
R.~Chalapathy and S.~Chawla, ``Deep learning for anomaly detection: A survey,''
  \emph{arXiv preprint arXiv:1901.03407}, 2019.

\bibitem{santhosh2020anomaly}
K.~K. Santhosh, D.~P. Dogra, and P.~P. Roy, ``Anomaly detection in road traffic
  using visual surveillance: A survey,'' \emph{ACM Computing Surveys (CSUR)},
  vol.~53, no.~6, pp. 1--26, 2020.

\bibitem{liu2018future}
W.~Liu, W.~Luo, D.~Lian, and S.~Gao, ``Future frame prediction for anomaly
  detection--a new baseline,'' in \emph{Proceedings of the IEEE conference on
  computer vision and pattern recognition}, 2018, pp. 6536--6545.

\bibitem{Ionescu_2019_CVPR}
R.~T. Ionescu, F.~S. Khan, M.-I. Georgescu, and L.~Shao, ``Object-centric
  auto-encoders and dummy anomalies for abnormal event detection in video,'' in
  \emph{Proceedings of the IEEE/CVF Conference on Computer Vision and Pattern
  Recognition (CVPR)}, June 2019.

\bibitem{xu2015learning}
D.~Xu, E.~Ricci, Y.~Yan, J.~Song, and N.~Sebe, ``Learning deep representations
  of appearance and motion for anomalous event detection,'' \emph{arXiv
  preprint arXiv:1510.01553}, 2015.

\bibitem{ma2015anomaly}
K.~Ma, M.~Doescher, and C.~Bodden, ``Anomaly detection in crowded scenes using
  dense trajectories,'' \emph{University of Wisconsin-Madison}, 2015.

\bibitem{lu2013abnormal}
C.~Lu, J.~Shi, and J.~Jia, ``Abnormal event detection at 150 fps in matlab,''
  in \emph{Proceedings of the IEEE international conference on computer
  vision}, 2013, pp. 2720--2727.

\bibitem{mahadevan2010anomaly}
V.~Mahadevan, W.~Li, V.~Bhalodia, and N.~Vasconcelos, ``Anomaly detection in
  crowded scenes,'' in \emph{2010 IEEE Computer Society Conference on Computer
  Vision and Pattern Recognition}.\hskip 1em plus 0.5em minus 0.4em\relax IEEE,
  2010, pp. 1975--1981.

\bibitem{adam2008robust}
A.~Adam, E.~Rivlin, I.~Shimshoni, and D.~Reinitz, ``Robust real-time unusual
  event detection using multiple fixed-location monitors,'' \emph{IEEE
  transactions on pattern analysis and machine intelligence}, vol.~30, no.~3,
  pp. 555--560, 2008.

\bibitem{teng2021viewpoint}
S.~Teng, S.~Zhang, Q.~Huang, and N.~Sebe, ``Viewpoint and scale consistency
  reinforcement for uav vehicle re-identification,'' \emph{International
  Journal of Computer Vision}, vol. 129, no.~3, pp. 719--735, 2021.

\bibitem{chriki2021deep}
A.~Chriki, H.~Touati, H.~Snoussi, and F.~Kamoun, ``Deep learning and
  handcrafted features for one-class anomaly detection in uav video,''
  \emph{Multimedia Tools and Applications}, vol.~80, no.~2, pp. 2599--2620,
  2021.

\bibitem{9219585}
------, ``Uav-based surveillance system: an anomaly detection approach,'' in
  \emph{2020 IEEE Symposium on Computers and Communications (ISCC)}, 2020, pp.
  1--6.

\bibitem{9341790}
I.~Bozcan and E.~Kayacan, ``Uav-adnet: Unsupervised anomaly detection using
  deep neural networks for aerial surveillance,'' in \emph{2020 IEEE/RSJ
  International Conference on Intelligent Robots and Systems (IROS)}, 2020, pp.
  1158--1164.

\bibitem{Ramachandra2020}
B.~Ramachandra, M.~Jones, and R.~R. Vatsavai, ``{A Survey of Single-Scene Video
  Anomaly Detection},'' \emph{IEEE Transactions on Pattern Analysis and Machine
  Intelligence}, vol. 8828, no.~c, pp. 1--20, 2020.

\bibitem{doshi2020any}
K.~Doshi and Y.~Yilmaz, ``Any-shot sequential anomaly detection in surveillance
  videos,'' in \emph{Proceedings of the IEEE/CVF Conference on Computer Vision
  and Pattern Recognition Workshops}, 2020, pp. 934--935.

\bibitem{xu2017detecting}
D.~Xu, Y.~Yan, E.~Ricci, and N.~Sebe, ``Detecting anomalous events in videos by
  learning deep representations of appearance and motion,'' \emph{Computer
  Vision and Image Understanding}, vol. 156, pp. 117--127, 2017.

\bibitem{zhao2011online}
B.~Zhao, L.~Fei-Fei, and E.~P. Xing, ``Online detection of unusual events in
  videos via dynamic sparse coding,'' in \emph{CVPR 2011}.\hskip 1em plus 0.5em
  minus 0.4em\relax IEEE, 2011, pp. 3313--3320.

\bibitem{cheng2015video}
K.-W. Cheng, Y.-T. Chen, and W.-H. Fang, ``Video anomaly detection and
  localization using hierarchical feature representation and gaussian process
  regression,'' in \emph{Proceedings of the IEEE Conference on Computer Vision
  and Pattern Recognition}, 2015, pp. 2909--2917.

\bibitem{raghavendra2006unusual}
R.~Raghavendra, A.~Bue, and M.~Cristani, ``Unusual crowd activity dataset of
  university of minnesota,'' 2006.

\bibitem{sultani2018real}
W.~Sultani, C.~Chen, and M.~Shah, ``Real-world anomaly detection in
  surveillance videos,'' in \emph{Proceedings of the IEEE conference on
  computer vision and pattern recognition}, 2018, pp. 6479--6488.

\bibitem{Ramachandra_2020_WACV}
B.~Ramachandra, M.~Jones, and R.~Vatsavai, ``Learning a distance function with
  a siamese network to localize anomalies in videos,'' in \emph{Proceedings of
  the IEEE/CVF Winter Conference on Applications of Computer Vision (WACV)},
  March 2020.

\bibitem{Sabokrou_2018_CVPR}
M.~Sabokrou, M.~Khalooei, M.~Fathy, and E.~Adeli, ``Adversarially learned
  one-class classifier for novelty detection,'' in \emph{Proceedings of the
  IEEE Conference on Computer Vision and Pattern Recognition (CVPR)}, June
  2018.

\bibitem{ionescu2019detecting}
R.~T. Ionescu, S.~Smeureanu, M.~Popescu, and B.~Alexe, ``Detecting abnormal
  events in video using narrowed normality clusters,'' in \emph{2019 IEEE
  Winter Conference on Applications of Computer Vision (WACV)}.\hskip 1em plus
  0.5em minus 0.4em\relax IEEE, 2019, pp. 1951--1960.

\bibitem{ravanbakhsh2018plug}
M.~Ravanbakhsh, M.~Nabi, H.~Mousavi, E.~Sangineto, and N.~Sebe, ``Plug-and-play
  cnn for crowd motion analysis: An application in abnormal event detection,''
  in \emph{2018 IEEE Winter Conference on Applications of Computer Vision
  (WACV)}.\hskip 1em plus 0.5em minus 0.4em\relax IEEE, 2018, pp. 1689--1698.

\bibitem{dalal2006human}
N.~Dalal, B.~Triggs, and C.~Schmid, ``Human detection using oriented histograms
  of flow and appearance,'' in \emph{European conference on computer
  vision}.\hskip 1em plus 0.5em minus 0.4em\relax Springer, 2006, pp. 428--441.

\bibitem{dalal2005histograms}
N.~Dalal and B.~Triggs, ``Histograms of oriented gradients for human
  detection,'' in \emph{2005 IEEE computer society conference on computer
  vision and pattern recognition (CVPR'05)}, vol.~1.\hskip 1em plus 0.5em minus
  0.4em\relax Ieee, 2005, pp. 886--893.

\bibitem{sabokrou2018deep}
M.~Sabokrou, M.~Fayyaz, M.~Fathy, Z.~Moayed, and R.~Klette, ``Deep-anomaly:
  Fully convolutional neural network for fast anomaly detection in crowded
  scenes,'' \emph{Computer Vision and Image Understanding}, vol. 172, pp.
  88--97, 2018.

\bibitem{sabokrou2017deep}
M.~Sabokrou, M.~Fayyaz, M.~Fathy, and R.~Klette, ``Deep-cascade: Cascading 3d
  deep neural networks for fast anomaly detection and localization in crowded
  scenes,'' \emph{IEEE Transactions on Image Processing}, vol.~26, no.~4, pp.
  1992--2004, 2017.

\bibitem{antic2015spatio}
B.~Anti{\'c} and B.~Ommer, ``Spatio-temporal video parsing for abnormality
  detection,'' \emph{arXiv preprint arXiv:1502.06235}, 2015.

\bibitem{kim2009observe}
J.~Kim and K.~Grauman, ``Observe locally, infer globally: a space-time mrf for
  detecting abnormal activities with incremental updates,'' in \emph{2009 IEEE
  conference on computer vision and pattern recognition}.\hskip 1em plus 0.5em
  minus 0.4em\relax IEEE, 2009, pp. 2921--2928.

\bibitem{benezeth2009abnormal}
Y.~Benezeth, P.-M. Jodoin, V.~Saligrama, and C.~Rosenberger, ``Abnormal events
  detection based on spatio-temporal co-occurences,'' in \emph{2009 IEEE
  conference on computer vision and pattern recognition}.\hskip 1em plus 0.5em
  minus 0.4em\relax IEEE, 2009, pp. 2458--2465.

\bibitem{kratz2009anomaly}
L.~Kratz and K.~Nishino, ``Anomaly detection in extremely crowded scenes using
  spatio-temporal motion pattern models,'' in \emph{2009 IEEE conference on
  computer vision and pattern recognition}.\hskip 1em plus 0.5em minus
  0.4em\relax IEEE, 2009, pp. 1446--1453.

\bibitem{feng2017learning}
Y.~Feng, Y.~Yuan, and X.~Lu, ``Learning deep event models for crowd anomaly
  detection,'' \emph{Neurocomputing}, vol. 219, pp. 548--556, 2017.

\bibitem{vu2019robust}
H.~Vu, T.~D. Nguyen, T.~Le, W.~Luo, and D.~Phung, ``Robust anomaly detection in
  videos using multilevel representations,'' in \emph{Proceedings of the AAAI
  Conference on Artificial Intelligence}, vol.~33, no.~01, 2019, pp.
  5216--5223.

\bibitem{chong2017abnormal}
Y.~S. Chong and Y.~H. Tay, ``Abnormal event detection in videos using
  spatiotemporal autoencoder,'' in \emph{International symposium on neural
  networks}.\hskip 1em plus 0.5em minus 0.4em\relax Springer, 2017, pp.
  189--196.

\bibitem{hasan2016learning}
M.~Hasan, J.~Choi, J.~Neumann, A.~K. Roy-Chowdhury, and L.~S. Davis, ``Learning
  temporal regularity in video sequences,'' in \emph{Proceedings of the IEEE
  conference on computer vision and pattern recognition}, 2016, pp. 733--742.

\bibitem{redmon2018yolov3}
J.~Redmon and A.~Farhadi, ``Yolov3: An incremental improvement,'' \emph{arXiv
  preprint arXiv:1804.02767}, 2018.

\bibitem{farneback2003two}
G.~Farneb{\"a}ck, ``Two-frame motion estimation based on polynomial
  expansion,'' in \emph{Scandinavian conference on Image analysis}.\hskip 1em
  plus 0.5em minus 0.4em\relax Springer, 2003, pp. 363--370.

\bibitem{sainju2014automated}
S.~Sainju, F.~M. Bui, and K.~A. Wahid, ``Automated bleeding detection in
  capsule endoscopy videos using statistical features and region growing,''
  \emph{Journal of medical systems}, vol.~38, no.~4, pp. 1--11, 2014.

\bibitem{cordts2016cityscapes}
M.~Cordts, M.~Omran, S.~Ramos, T.~Rehfeld, M.~Enzweiler, R.~Benenson,
  U.~Franke, S.~Roth, and B.~Schiele, ``The cityscapes dataset for semantic
  urban scene understanding,'' in \emph{Proceedings of the IEEE conference on
  computer vision and pattern recognition}, 2016, pp. 3213--3223.

\bibitem{girisha2021uvid}
S.~Girisha, U.~Verma, M.~M. Pai, and R.~M. Pai, ``Uvid-net: Enhanced semantic
  segmentation of uav aerial videos by embedding temporal information,''
  \emph{IEEE Journal of Selected Topics in Applied Earth Observations and
  Remote Sensing}, vol.~14, pp. 4115--4127, 2021.

\bibitem{cong2011sparse}
Y.~Cong, J.~Yuan, and J.~Liu, ``Sparse reconstruction cost for abnormal event
  detection,'' in \emph{CVPR 2011}.\hskip 1em plus 0.5em minus 0.4em\relax
  IEEE, 2011, pp. 3449--3456.

\bibitem{liu2018classifier}
Y.~Liu, C.-L. Li, and B.~P{\'o}czos, ``Classifier two sample test for video
  anomaly detections.'' in \emph{BMVC}, 2018, p.~71.

\bibitem{astrid2021synthetic}
M.~Astrid, M.~Z. Zaheer, and S.-I. Lee, ``Synthetic temporal anomaly guided
  end-to-end video anomaly detection,'' in \emph{Proceedings of the IEEE/CVF
  International Conference on Computer Vision}, 2021, pp. 207--214.

\bibitem{li2021two}
T.~Li, X.~Chen, F.~Zhu, Z.~Zhang, and H.~Yan, ``Two-stream deep
  spatial-temporal auto-encoder for surveillance video abnormal event
  detection,'' \emph{Neurocomputing}, vol. 439, pp. 256--270, 2021.

\end{thebibliography}
%

%

\begin{IEEEbiography}[{\includegraphics[width=1in,height=1in,clip]{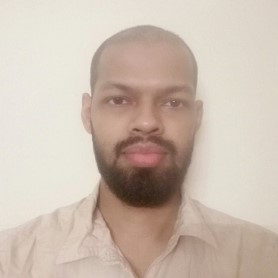}}]{Girisha S}
is pursuing PhD in Computer vision and Deep learning at Manipal Institute of Technology. He has completed his masters degree in computer science and engineering from NMAMIT, Nitte, India and B.E degree from srinivas school of engineering affiliated to VTU, Belgaum. His area of interest is image segmentation, object detection and deep learning for computer vision.
\end{IEEEbiography}
\vfill
\begin{IEEEbiography}[{\includegraphics[width=1in,height=1.25in,clip,keepaspectratio]{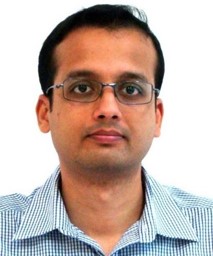}}]{DR. UJJWAL VERMA }
received the Ph.D. degree from Télécom ParisTech, University of Paris Saclay, Paris, France in Image Processing, and M.S. (Research) degree from IMT Atlantique (France) in Signal and Image Processing. Currently, he is an Associate Professor and Head in the Department of Electronics and Communication Engineering, Manipal Institute of Technology, Bengaluru, India. His research interests include variational methods in image segmentation, action recognition, and deep learning methods for scene understanding. He is a recipient of "ISCA Young Scientist Award 2017-18" by Indian Science Congress Association (ISCA), a professional body under the Department of Science and Technology, Government of India. He is a senior member of IEEE and is currently Co-Chair, Working Group on Machine/Deep Learning for Image Analysis, Technical Committee on Image Analysis and Data Fusion (IADF), IEEE Geoscience and Remote Sensing Society (GRSS). He is also a member of Executive committee of IEEE GRSS Bangalore Chapter. 
\end{IEEEbiography}
\begin{IEEEbiography}[{\includegraphics[width=1in,height=1.25in,clip,keepaspectratio]{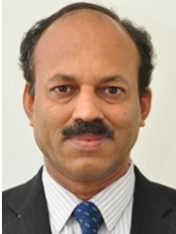}}]{DR. MANOHARA M. M. PAI}
holds Ph.D. in Computer Science and Engineering and is the Professor in the Department of Information and Communication Technology at Manipal Institute of Technology, Manipal Academy of Higher Education, Manipal, India for the last 30 years. He holds 7 patents to his credit and has published 86 papers in National and International Journals/ Conference proceedings. He has published two books, guided 6 PhDs and 85 Master thesis. His areas of interest includes Data Analytics, Cloud computing, IoT, Computer Networks, Mobile Computing, Scalable Video Coding, Robot Motion Planning. He is Senior Member of IEEE, Life member of ISTE and Life member of Systems Society of India. He is the Principal Investigator for multiple Industry/Govt. research projects. He is the Executive Committee member of IEEE Mangalore Subsection and past Chair of IEEE Mangalore Subsection (2019).
\end{IEEEbiography}
\begin{IEEEbiography}[{\includegraphics[width=1in,height=1.25in,clip,keepaspectratio]{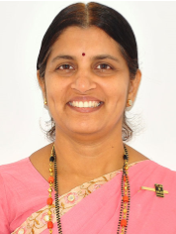}}]{DR. RADHIKA M. PAI}
is a Professor in the Department of Information and Communication Technology at Manipal Institute of Technology, Manipal Academy of Higher Education, Manipal, India. She obtained her Ph.D from National Institute of Technology Karnataka, Surathkal, India and  is a recipient of National Doctoral fellowship from AICTE, Govt. of India. She has a teaching and research experience of above 30 years. She has published 63 papers in National/International Journals/Conferences and has guided 3 PhDs and several Master thesis. Her areas of interest include Data Mining, Big Data Analytics, Character Recognition, Sensor networks and e-learning. She is a senior member of IEEE and Executive Committee member of IEEE Mangalore Subsection.
\end{IEEEbiography}

\vfill




\end{document}